\documentclass[lettersize,journal,compsoc]{IEEEtran}

\ifCLASSOPTIONcompsoc
  \usepackage[nocompress]{cite}
\else
  \usepackage{cite}
\fi

\ifCLASSINFOpdf
\else
\fi

\usepackage{amsmath}
\usepackage{array}
\usepackage{etoolbox}
\usepackage{textcomp}
\usepackage[nocompress]{cite}
\usepackage{stfloats}
\usepackage{url}
\usepackage{verbatim}
\usepackage{graphicx}
\usepackage{mathrsfs}
\usepackage{booktabs}       %
\usepackage{nicefrac}       %
\usepackage{microtype}      %
\usepackage{float}
\usepackage{multirow}
\usepackage{makecell}
\usepackage{pifont}
\usepackage{listings}
\usepackage{enumitem}
\usepackage{color, colortbl}
\usepackage{wrapfig}
\usepackage[square,sort,comma,numbers]{natbib}
\newcolumntype{x}[1]{>{\centering\arraybackslash}p{#1pt}}
\newcolumntype{y}[1]{>{\raggedright\arraybackslash}p{#1pt}}
\newcolumntype{z}[1]{>{\raggedleft\arraybackslash}p{#1pt}}
\newcommand{\tablestyle}[2]{\setlength{\tabcolsep}{#1}\renewcommand{\arraystretch}{#2}\centering\footnotesize}
\usepackage{changepage,threeparttable} %

\usepackage{array}
\usepackage{epsfig}
\usepackage{graphicx}
\usepackage{float}
\usepackage{wrapfig}
\usepackage{amsmath,amssymb,amsthm}
\usepackage{algorithm,algorithmicx,algpseudocode}
\usepackage{bm,xspace}
\usepackage{comment}
\usepackage{multirow}
\usepackage{balance}
\usepackage{url}
\usepackage{booktabs}
\usepackage{etoolbox,siunitx}
\usepackage{calc}
\usepackage{pifont,hologo}
\usepackage{color}
\usepackage{adjustbox}
\usepackage{amsmath}
\usepackage{enumitem}
\usepackage{subcaption}
\usepackage{titlesec}
\usepackage{bbding}  %
\usepackage[normalem]{ulem}  %
\PassOptionsToPackage{table}{xcolor}
\usepackage{colortbl}
\usepackage[table]{xcolor}
\usepackage[pagebackref,breaklinks,colorlinks,linkcolor=red,citecolor=blue]{hyperref}

\newcommand\revise[1]{\textcolor{black}{#1}}
\newcommand\minorrevise[1]{\textcolor{black}{#1}}
\newcommand\revisenew[1]{\textcolor{black}{#1}}

\definecolor{blue}{HTML}{85a67d}
\definecolor{red}{HTML}{9f836f}
\definecolor{orange}{HTML}{cc7700}
\definecolor{gray}{HTML}{efefef}
\definecolor{darkgreen}{HTML}{228B22}
\definecolor{darkgray}{HTML}{757575}

\newcommand{\figref}[1]{Fig.~\ref{#1}}
\newcommand{\tabref}[1]{Tab.~\ref{#1}}
\newcommand{\secref}[1]{Sec.~\ref{#1}}
\renewcommand{\eqref}[1]{Eq.~\ref{#1}}

\DeclareMathSymbol{@}{\mathord}{letters}{"3B}

\makeatletter
\DeclareRobustCommand\onedot{\futurelet\@let@token\@onedot}
\def\@onedot{\ifx\@let@token.\else.\null\fi\xspace}
\def\eg{\emph{e.g}\onedot} 
\def\ie{\emph{i.e}\onedot} 
 
\def\etc{\emph{etc}\onedot}

\newcommand*{\Rom}[1]{\expandafter\@slowromancap\romannumeral #1@}
\newcommand*{\rom}[1]{\expandafter\romannumeral #1}

\def\1{\bm{1}}

\def\rvd{{\mathbf{d}}}

\def\rvf{{\mathbf{f}}}

\def\rvh{{\mathbf{h}}}

\def\rvl{{\mathbf{l}}}

\def\rvn{{\mathbf{n}}}
\def\rvo{{\mathbf{o}}}
\def\rvp{{\mathbf{p}}}

\def\rvr{{\mathbf{r}}}

\def\rvx{{\mathbf{x}}}

\def\ervr{{\textnormal{r}}}

\def\rmP{{\mathbf{P}}}

\def\vg{{\bm{g}}}

\def\evg{{g}}

\def\mI{{\bm{I}}}

\def\tB{{\mathcal{B}}}
\def\tC{{\mathcal{C}}}

\def\tF{{\mathcal{F}}}

\def\tV{{\mathcal{V}}}

\def\sI{{\mathcal{I}}}

\def\stF{{\mathscr{F}}}

\def\stX{{\mathscr{X}}}

\def\fG{{\mathbb{\mathcal{G}}}}

\def\fT{{\mathbb{\mathcal{T}}}}

\newcommand{\Ls}{\mathcal{L}}
\newcommand{\R}{\mathbb{R}}
\newcommand{\N}{\mathbb{N}}

\newcommand{\sigmoid}{\sigma}

\newcommand{\normlone}{L^1}

\let\originalleft\left
\let\originalright\right
\renewcommand{\left}{\mathopen{}\mathclose\bgroup\originalleft}
\renewcommand{\right}{\aftergroup\egroup\originalright}

\newcommand\up[1]{\textcolor{red}{$^{\uparrow{#1}}$}}

\usepackage{hyperref}

\def\sexyname{{PonderV2}\xspace}
\usepackage[table]{xcolor}

\begin{document}
\definecolor{mgreen}{RGB}{1,150,74}
\title{\sexyname: Improved 3D Representation with \\A Universal Pre-training Paradigm}

\author{Haoyi Zhu$^{1,2*}$\thanks{$^*$Equal contribution.}, 
Honghui Yang$^{2,4*}$, 
Xiaoyang Wu$^{2,3*}$, 
Di Huang$^{2*}$,
Sha Zhang$^{1,2}$, 
Xianglong He$^{2}$, \\
Hengshuang Zhao$^{3}$, 
Chunhua Shen$^{4}$, 
Yu Qiao$^{2}$, 
Tong He$^{2\dag}$\thanks{$^\dag$Corresponding author. Email: {\tt\footnotesize \href{mailto:hetong@pjlab.org.cn}{hetong@pjlab.org.cn}}.
}, 
Wanli Ouyang$^{2}$\\

$^1$University of Science and Technology of China\quad
$^2$Shanghai Artificial Intelligence Laboratory\\
$^3$The University of Hong Kong\quad
$^4$Zhejiang University

}

\markboth{Journal of \LaTeX\ Class Files,~Vol.~14, No.~8, August~2021}%
{Shell \MakeLowercase{\textit{et al.}}: A Sample Article Using IEEEtran.cls for IEEE Journals}

\IEEEtitleabstractindextext{

\begin{abstract}

In contrast to numerous NLP and 2D vision foundational models, \minorrevise{training} a 3D foundational model poses considerably greater challenges. This is primarily due to the inherent data variability and diversity of downstream tasks. In this paper, we introduce a novel universal 3D pre-training framework designed to facilitate the acquisition \minorrevise{of efficient 3D representations}.
Considering that informative 3D features should encode rich geometry and appearance cues that can be utilized to render realistic images, we propose to learn 3D representations by differentiable neural rendering. \minorrevise{We train a 3D backbone with a volumetric neural renderer} by comparing the rendered with the real images. 
\revise{Notably, our pre-trained encoder can be seamlessly applied to various downstream tasks.}
\minorrevise{These tasks include semantic challenges like 3D detection and segmentation, which involve scene understanding, and non-semantic tasks like 3D reconstruction and image synthesis, which focus on geometry and visuals. They span both indoor and outdoor scenarios.}
\minorrevise{We also} illustrate the capability of pre-training a 2D backbone using the proposed methodology, surpassing conventional pre-training methods by a large margin. For the first time, PonderV2 achieves state-of-the-art performance on 11 indoor and outdoor benchmarks, implying its effectiveness. Code and models are available at \url{https://github.com/OpenGVLab/PonderV2}.
\end{abstract}


}
\maketitle

\IEEEdisplaynontitleabstractindextext

\IEEEpeerreviewmaketitle

\IEEEraisesectionheading{\section{Introduction}\label{sec:introduction}}

\IEEEPARstart{P}{r}\revise{\minorrevise{e-trained models}} hold immense importance and have seen extensive development across various fields, including NLP~\cite{radford2018improving,radford2019language,brown2020language,touvron2023llama,touvron2023llama2,chowdhery2022palm,2023internlm}, 2D computer vision~\cite{kirillov2023segment,rombach2022high,wang2023internimage,wang2022internvideo,he2022masked}, multimodal domains~\cite{radford2021learning,yu2022coca,zhang2023internlm,chen2022pali,alayrac2022flamingo,li2022blip,li2023scaling} and embodied AI~\cite{driess2023palme,fan2022minedojo,brohan2022rt,brohan2023rt,fang2023rh20t}. 
\revise{Pretraining is significant as it provides models with a strong foundation of learned features, enabling better generalization and faster convergence across a wide range of downstream tasks, thus significantly saving computational resources and time.} 
Although promising, developing such a model in 3D is more challenging. Firstly, the diversity of 3D representations, such as \revise{point clouds and multi-view images}, introduces complexities in designing a universal pre-training approach. 
\minorrevise{Moreover, the predominance of empty space within 3D data and the variability in both data format and quality caused by sensor placement and occlusions present unique obstacles in acquiring generalizable features.}

\begin{figure}[!htb]
	\centering
	\includegraphics[width=0.98\linewidth]{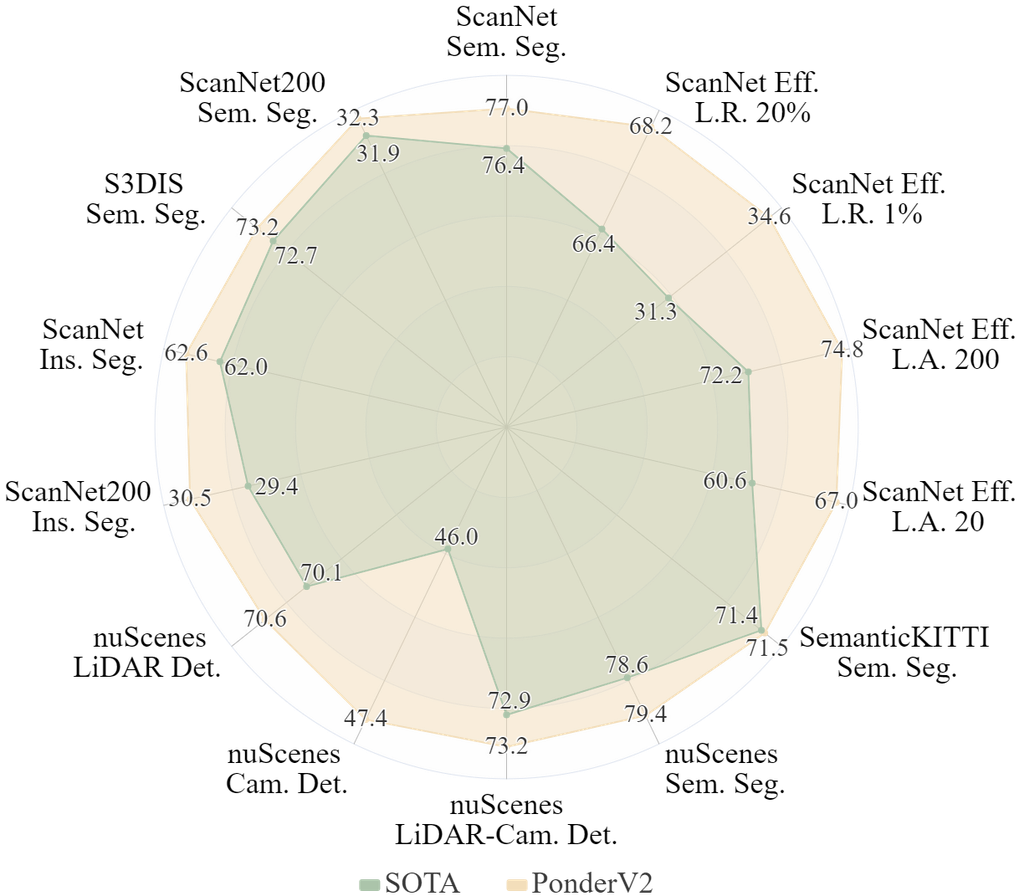}
	\caption{\textbf{The radar chart of \sexyname,} showing its effectiveness on over 10 benchmarks in both indoor and outdoor scenarios. Abbreviations: sem. for semantic, ins. for instance, seg. for segmentation, eff. for efficient, L.R. for limited reconstructions, L.A. for limited annotations, obj. for object, rec. for reconstruction, cam. for camera, det. for detection. The SOTA in the figure denotes the state-of-the-art performance with the same backbone as ours on validation sets. \revise{The units for different tasks are normalized based on the minimum and maximum performance for each task.}}
	\label{fig:radar_all}
	\vspace{-2em}
\end{figure}

Previous pre-training methods for obtaining effective 3D representation can be roughly categorized into two groups: contrast-based~\cite{xie2020pointcontrast, hou2021exploring, jiang2021guided, huang2021spatio, chen20224dconstrast, rao2021randomrooms, zhang_depth_contrast} and masked autoencoding-based (or completion-based)~\cite{wang2021occo, yu2022pointbert, yan2022implicit, pang2022masked, liu2022maskpoint, zhang2022pointm2ae, min2022voxelmae}.
\revise{Contrast-based methods are designed to maintain invariant representation under different transformations, which requires informative samples during training.} In the 2D domain, it is addressed by (1) introducing efficient positive/negative sampling methods, (2) using a large batch size and storing representative samples, and (3) applying various data augmentation policies.
Inspired by these works, many works~\cite{xie2020pointcontrast, hou2021exploring, jiang2021guided, huang2021spatio, chen20224dconstrast, rao2021randomrooms, zhang_depth_contrast} are proposed to learn geometry-invariant features on 3D point clouds. 

Methods using masked autoencoding are another line of research for 3D representation learning, which utilizes a pre-training task of reconstructing the masked point cloud based on partial observations. 
By maintaining a high masking ratio, this simple task encourages the model to learn a holistic understanding of the input beyond low-level statistics. 
\revise{Although masked autoencoders have been successfully applied in 2D images~\cite{he2022masked} and videos~\cite{fei2022videomae, tong2022videomae}, their application to 3D scenes remains an open challenge \revisenew{due to the irregularity and low-density format} of point cloud.}

\begin{figure}[!tb]
	\centering
	\includegraphics[width=0.98\linewidth]{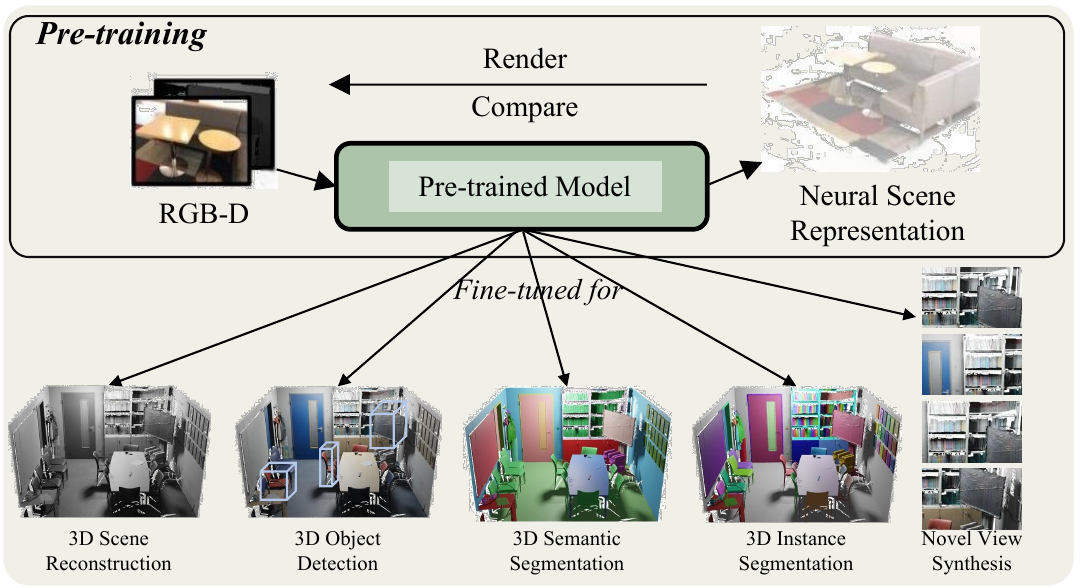}
	\caption{The proposed unified 3D pre-training approach, 
 termed 
 \textbf{\sexyname}, is directly trained with RGB-D rendered image supervision, and can be used for \revise{various 3D downstream applications, \eg, 3D object detection, 3D semantic segmentation, 3D scene reconstruction, and image synthesis}.}
	\label{fig:teaser_tasks}
	\vspace{-1.2em}
\end{figure}

Different from the methods above, we propose \textbf{po}int cloud pre-training via neural re\textbf{nder}ing (PonderV2), an extension of our preliminary work published at conference ICCV 2023~\cite{huang2023ponder}, as shown in \figref{fig:teaser_tasks}.
Our motivation is that \revise{neural rendering can be leveraged to ensure point cloud features encapsulate rich geometry and appearance cues}. \revise{Pre-training with rendering creates an effective implicit 3D representation. Our method accommodates a wide variety of supervision signals, making it versatile and robust.}

\revise{Specifically, 3D data formats, such as multi-view images or point clouds, are processed through an encoder, then a decoder, to generate RGB or depth images using differentiable rendering techniques.}
The network is trained to minimize the difference between rendered and observed 2D images. In doing so, our method implicitly encodes 3D space, facilitating the reconstruction of continuous 3D shape structures and the intricate appearance characteristics of their 2D projections. 
Furthermore, given the 2D nature of the loss terms, it provides the flexibility to employ diverse signals for supervision, including RGB images, depth images, and semantic knowledge from 2D foundation models.

\revise{To validate the effectiveness of our method, we conducted extensive experiments across six indoor and outdoor \minorrevise{benchmarks}}, including high-level tasks such as segmentation and detection, as well as low-level tasks like image synthesis and scene/object reconstruction. We achieved state-of-the-art performance on more than 11 benchmarks. Part of \sexyname's validation set performance, compared to baselines and state-of-the-art methods with the same backbone, is shown in \figref{fig:radar_all}.
\minorrevise{The results} indicate the effectiveness of the proposed universal methodology. Specifically, we first evaluate \sexyname on different backbones across various popular indoor benchmarks with multi-frame RGB-D images as inputs, demonstrating its flexibility. Furthermore, we pre-train a single backbone for various downstream tasks, namely SparseUNet~\cite{choy20194d}, which processes whole-scene point clouds and remarkably surpasses state-of-the-art methods with the same backbone on various indoor 3D benchmarks. For example, \sexyname reaches $77.0$ {\em val} mIoU on ScanNet semantic segmentation benchmark and ranks $1$st on ScanNet benchmark with a {\em test} mIoU of $78.5$. \sexyname also ranks $1$st on ScanNet200 semantic segmentation benchmark with a {\em test} mIoU of $34.6$. 
\minorrevise{Note that after pre-training, we initialize the encoder weights with the pre-trained backbone and do fine-tuning for downstream tasks.}
We conduct extensive experiments in outdoor autonomous driving scenarios, reaching state-of-the-art validation performance as well. For example, we achieved 73.2 NDS for 3D detection and 79.4 mIoU for 3D segmentation on the nuScenes validation set, 3.0 and 6.1 higher than the baseline, respectively.  \revise{For outdoor scenarios, the BEV volume is also pre-trained and loaded.} The promising results demonstrate the efficacy of \sexyname.

\revise{In summary, our main contributions are the following:}
\begin{itemize}
    \item \revise{\minorrevise{We propose to use differentiable neural rendering as a novel pre-training paradigm tailored for 3D vision tasks. 
    Our novelty lies not in neural rendering itself, but in how it is employed for representation learning in 3D vision downstream tasks.}}

    \item \revise{Our approach is highly effective in acquiring efficient 3D representations, adept at encoding detailed geometric and visual information through neural rendering. This flexible framework is applicable across various modalities, covering both indoor and outdoor environments, including but not limited to point clouds and multi-view images.}

    \item The proposed methodology reaches state-of-the-art performance \minorrevise{on many popular indoor and outdoor benchmarks}, and is flexible to integrate into various backbones. Besides high-level perception tasks, \sexyname can also boost low-level tasks such as image synthesis, scene and object reconstruction, \etc. The effectiveness and flexibility of \sexyname showcase the potential to pave the way for a 3D foundation model.

\end{itemize}

\section{Related Works}

\noindent \textbf{Self-supervised learning in point clouds.}
\revise{
Current methods can be broadly categorized into contrast-based and masked autoencoding approaches.} Inspired by 2D domain works~\cite{he2020momentum, chen2020simple}, PointContrast~\cite{xie2020pointcontrast} pioneers 3D contrastive learning, encouraging invariant 3D representations. Some works~\cite{hou2021exploring, jiang2021guided, huang2021spatio, chen20224dconstrast, rao2021randomrooms, zhang_depth_contrast, wu2023masked} devise new sampling strategies or data augmentations.
Masked Autoencoding methods~\cite{yu2022pointbert, yan2022implicit, pang2022masked, liu2022maskpoint, zhang2022pointm2ae, min2022voxelmae}, inspired by MAE~\cite{he2022masked}, include PointMAE~\cite{pang2022masked} using Chamfer Distance, PointM2AE~\cite{zhang2022pointm2ae} with a multiscale strategy, VoxelMAE~\cite{min2022voxelmae} recovering geometry, and GD-MAE~\cite{yang2023gd-mae} with a generative decoder. ALSO~\cite{boulch2023also} focuses on surface reconstruction as a pretext task.
Different from the above pipelines, we propose a novel framework for point cloud pre-training via neural rendering, applicable to both image- and point-based models.

\noindent \textbf{Representation learning on images.} \revise{Representation learning has been well-developed in the 2D domain~\citep{he2022masked, bachmann2022multimae, bao2021beit, he2020momentum, chen2020improved, tong2022videomae}.} %
Contrastive-based methods, such as MoCo~\citep{he2020momentum,chen2020improved}, learn images' representations by discriminating the similarities between different augmented samples.
MAE-based methods~\cite{gao2022convmae,tian2023designing}, obtain the promising generalization ability by recovering the masked patches.
Within the realm of 3D applications, models pre-trained on ImageNet~\citep{deng2009imagenet} are widely utilized in image-related tasks~\citep{liu2022petr, liu2023sparsebev, liang2022bevfusion, li2022uvtr, yan2023cmt, yang2022graphrcnn}.
For example, to compensate for the insufficiency of 3D priors in tasks like 3D object detection, depth estimation~\citep{park2021dd3d} and monocular 3D detection~\citep{wang2021fcos3d} are usually used as extra pre-training techniques.

\noindent \textbf{Neural rendering.} 
\minorrevise{Neural Rendering uses neural networks to render images} from 3D scene representation in a differentiable way.
NeRF~\cite{ben2020nerf} first represents the scene as the neural radiance field and renders the images via volume rendering.
Based on NeRF, there are a series of works~\cite{thomas2022instantngp, yu2021plenoctrees, wang2021neus, oechsle2021unisurf, Yu20arxiv_pixelNeRF, wang2021ibrnet, Zhang20arxiv_nerf++, reiser2021kilonerf, barron2021mip, MIR-2022-06-204, huang2023ponder} trying to improve the NeRF representation, including \revise{accelerating} NeRF training, \minorrevise{boosting the quality of geometry}, and so on.
Another type of neural rendering leverages neural point clouds as the scene representation. 
\cite{aliev2020neural, rakhimov2022npbg++} take \revise{points'} locations and corresponding descriptors as input, \minorrevise{rasterizing the points with z-buffer, and using a rendering network} to get the final image. 
Later work such as PointNeRF~\cite{xv2022pointnerf} and X-NeRF~\cite{zhu2022x} render realistic images from neural point representation using a NeRF-like rendering process.

\section{Neural Rendering as a Universal Pre-training Paradiagm}
\label{sec:main_method}
\begin{figure*}[!t]
	\centering
	\includegraphics[width=0.9\textwidth]{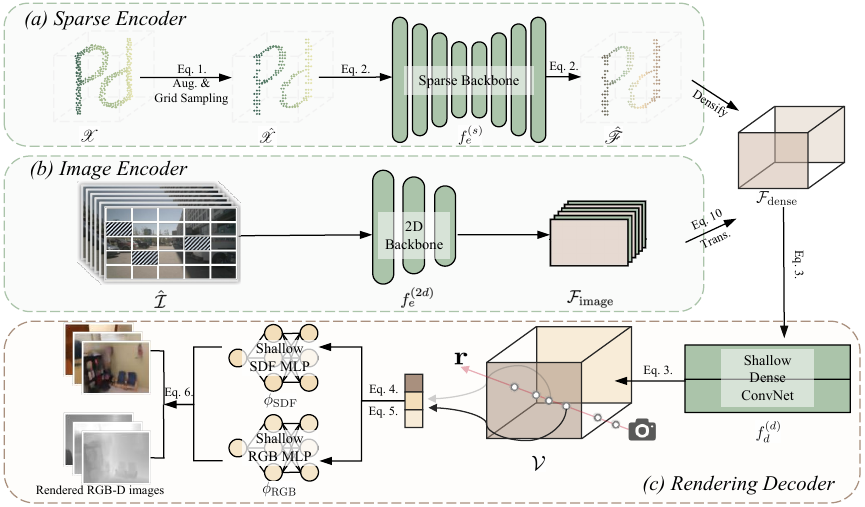}
	\caption{\textbf{The overall pipeline of \sexyname with \minorrevise{different modalities as input.}} \revise{\minorrevise{(a) \textit{Sparse point cloud as input.} A raw point cloud can be constructed from multi-frame RGB-D images, scene scans, or LiDAR.} We apply augmentations like masking and grid sampling to create a quantized sparse tensor. A sparse backbone extracts features, serving as the encoder for pre-training and as the pre-trained weight for fine-tuning. These sparse features are then densified to a feature volume. (b) \minorrevise{\textit{Multi-view images as input.} The image encoder takes augmented multi-view images as inputs, giving out the multi-view features. The features are then zero-padded and transformed to obtain the 3D dense feature volume. (c) \textit{\revisenew{Rendering} decoder as a unified pre-training paradiagm.}} A shallow dense convolutional network processes the dense feature volume to produce a dense feature volume. The rendering decoder queries this volume and uses shallow MLPs to output each point's SDF and color. Finally, these outputs are integrated to \minorrevise{render RGB-D images}, which are supervised by ground truth.}}
	\label{fig:overall_stru}
	\vspace{-1.2em}
\end{figure*}

In this section, we present the details of our methodology. We first give an overview of our pipeline in \secref{sec:method_pipeline_overview}, which is visualized in \figref{fig:overall_stru}. Then, we detail some specific differences and trials for indoor and outdoor scenarios due to the different input data types and settings in \secref{sec:method_indoor_scenario} and \secref{sec:method_outdoor_scenario}.

\subsection{Universal Pipeline Overview}
\label{sec:method_pipeline_overview}
\revise{We first apply our method to the 3D point cloud modality and then provide a detailed description of how our approach extends to other 3D modalities, such as multi-view images.}
As shown in \figref{fig:overall_stru}, the input of our pipeline is \revise{a sparse point cloud} $\stX = \{\tC_{\mathrm{in}}, \tF_{\mathrm{in}}\}$ comprising a set of $n$ coordinates $\tC_{\mathrm{in}} \in \mathbb{R}^{n \times 3}$ and their corresponding $ch_{in}$ channel features $\tF_{\mathrm{in}} \in \R^{n \times ch_{in}}$ which may include attributes such as colors or intensities. These point clouds can be generated from RGB-D images, scans, or LiDAR data. Before diving into our backbone, we first apply augmentations to the input data and quantize it using a specific grid size $\vg=[\evg_x, \evg_y, \evg_z] \in \mathbb{R}^3$. This process can be expressed as:
\begin{equation}
\hat{\stX} = \fG(\fT(\stX), \vg)) = \{\hat{\tC}_{\mathrm{in}}, \hat{\tF}_{\mathrm{in}}\} \text{,}
\end{equation}
where $\fG(\cdot, \vg)$ is a grid sampling function designed to ensure each grid has only one point sampled.
$\fT(\cdot)$ denotes the augmentation function, and $\hat{\stX}$ is the sampled points.

Then, we feed $\hat{\stX}$ into a sparse backbone $f_e^{(s)}(\cdot)$, serving as our pre-training encoder.
The outputs are obtained by:
\begin{equation}
\hat{\stF} = f_e^{(s)}(\hat{\stX}) = \{\hat{\tC}_{\mathrm{out}}, \hat{\tF}_{\mathrm{out}}\},
\end{equation}
where $\hat{\tC}_{\mathrm{out}}$ and $\hat{\tF}_{\mathrm{out}}$ are the coordinates and features of the sparse outputs, respectively.
To make the sparse features compatible with our volume-based decoder, we encode them into a volumetric representation by a densification process.
Specifically, we first discretize the 3D space at a resolution of $l_x \times l_y \times l_z $ voxel grids.
Subsequently, the sparse features that fall into the same voxel are aggregated by applying average pooling based on their corresponding sparse coordinates.
The aggregation will result in the dense volume features $\tF_\mathrm{dense} \in \R^{l_x \times l_y \times l_z \times ch_{out}}$, where empty voxel grids are padded with zeros.
A shallow dense convolutional network $f_d^{(d)}(\cdot)$ is then applied to obtain the enhanced 3D feature volume $\tV  \in \R^{l_x \times l_y \times l_z \times ch_{vol}}$, which can be expressed as:
\begin{equation}
    \tV = f_d^{(d)}(\tF_\mathrm{dense})
\end{equation}
Given the dense 3D volume $\tV$, we make a novel use of differentiable volume rendering to reconstruct the projected color images and depth images as the pretext task.
Inspired by \cite{wang2021neus}, we represent a scene as an implicit signed distance function (SDF) field to be capable of representing high-quality geometry details.
Specifically, given a camera pose $\rmP$ and sampled pixels $\rvx$, we shoot rays $\rvr$ from the camera's projection center $\rvo$ in direction $\rvd$ towards the pixels, which can be derived from its intrinsics and extrinsics. 
Along each ray, we sample $D$ points $\{\rvp_j = \rvo + t_j \cdot \rvd \mid j=1, ..., D \land 0\leq t_j < t_{j+1} \}$, where $t_j$ is the distance from each point to camera center, and query each point's 3D feature $\rvf_j$ from $\tV$ by trilinear interpolation.
A SDF value $s_j$ is predicted for each point $\rvp_j$ using an shallow MLP $\phi_\mathrm{SDF}$:
\begin{equation}
    s_j=\phi_\mathrm{SDF}(\rvp_j, \rvf_j) \text{,}
\end{equation}
To determine the color value, our approach draws inspiration from \citep{oechsle2021unisurf} and conditions the color field on the surface normal $\rvn_j$ (\ie, the gradient of the SDF value at ray point $\rvp_j$) together with a geometry feature vector $\rvh_i$ derived from $\phi_\mathrm{SDF}$.
This can yield a color representation:
\begin{equation}
    c_j=\phi_\mathrm{RGB}(\rvp_j, \rvf_j, \rvd_i, \rvn_j, \rvh_j) \text{,}
\end{equation}
where $\phi_\mathrm{RGB}$ is parameterized by another shallow MLP.
Subsequently, we render 2D colors $\hat{C}(\rvr)$ and depths $\hat{D}(\rvr)$ by integrating predicted colors and sampled depths along rays $\rvr$ using the following equations:
\begin{equation}
\label{eq:render}
	\hat{C}(\rvr)=\sum_{j=1}^D{w_j c_j}, \quad  \hat{D}(\rvr)=\sum_{j=1}^D{w_j t_j},
\end{equation}
The weight $w_j$ in these equations is an unbiased, occlusion-aware factor, as illustrated in \cite{wang2021neus}, and is computed as $w_j=T_j\alpha_j$. Here, $T_j = \prod_{k=1}^{j-1}(1-\alpha_k)$ represents the accumulated transmittance, while $\alpha_j$ is the opacity value computed by:
\begin{equation}
	\alpha_j=\max \left(\frac{\sigmoid_s\left(s_j\right)-\sigmoid_s\left(s_{j+1}\right)}{\sigmoid_s\left(s_j\right)}, 0\right),
\end{equation}
where $\sigmoid_s(x)=(1+e^{-sx})^{-1}$ is the $\operatorname{Sigmoid}$ function modulated by a learnable parameter $s$.

Finally, our optimization target is to minimize the $\normlone$ reconstruction loss on rendered 2D-pixel space with a $\lambda_C$ and a $\lambda_D$ factor adjusting the loss weights, namely:
\begin{equation}
\label{loss:render}
    \revise{\Ls_\mathrm{render}} = \frac{1}{| \rvr |} \sum_{\ervr \in \rvr} \lambda_C \cdot\Vert \hat{C}(\ervr) - C(\ervr) \Vert + \lambda_D \cdot \Vert \hat{D}(\ervr) - D(\ervr) \Vert
\end{equation}

\subsection{Indoor Scenario}
\label{sec:method_indoor_scenario}
\revise{\minorrevise{In the indoor scenario, we incorporate} the Eikonal loss $\mathcal{L}_{\text{eikonal}}$, near-surface SDF loss $\mathcal{L}_{\text{sdf}}$, and free space loss $\mathcal{L}_{\text{free}}$, which are commonly applied in neural surface reconstruction. Detailed definitions are provided in \textcolor{red}{Appendix}~\ref{appx:loss}.
}
\revise{In the conference paper~\cite{huang2023ponder}, the loss is defined as:
\begin{align}
\label{loss:indoor_rgbd}
\mathcal{L}_{\text{indoor}} &= \mathcal{L}_{\text{render}} + \lambda_{\text{eikonal}} \cdot \mathcal{L}_{\text{eikonal}} + \lambda_{\text{sdf}} \cdot \mathcal{L}_{\text{sdf}} + \lambda_{\text{free}} \cdot \mathcal{L}_{\text{free}}
\end{align}
where $\lambda_{\text{eikonal}}$, $\lambda_{\text{sdf}}$, and $\lambda_{\text{free}}$ are the corresponding loss weights.
}

\revise{
We also explore applying an additional semantic rendering decoder for indoor scenes. 
}
We use an additional shallow MLP, denoted as $\phi_\mathrm{SEMANTIC}$, to predict semantic features for each query point:
\begin{equation}
    \rvl_j = \phi_\mathrm{SEMANTIC}(\rvp_j, \rvf_j, \rvn_j, \rvh_j) \text{.}
\end{equation}

These semantic features can be projected onto a 2D canvas $\hat{F}$ using a similar weighting scheme as described in \eqref{eq:render}. For supervision, we utilize the CLIP~\cite{radford2021learning} features of each pixel's text label which is a readily available attribute in most indoor datasets. \revise{The supervision can also derived unsupervised from 2D foundation vision models, distilling knowledge from 2D foundation models into 3D backbones. For example, we can render the feature map of CLIP's visual encoder.} \revise{Let's denote the loss weight of semantic feature map rendering as $\lambda_F$. Now the total loss for the indoor scenario is defined as:
\begin{align}
\label{loss:indoor}
\mathcal{L}_{\text{indoor}} &= \mathcal{L}_{\text{render}} + \frac{1}{| \rvr |} \sum_{\ervr \in \rvr} \lambda_F \cdot\Vert \hat{F}(\ervr) - F(\ervr) \Vert \nonumber \\
&+ \lambda_{\text{eikonal}} \cdot \mathcal{L}_{\text{eikonal}} + \lambda_{\text{sdf}} \cdot \mathcal{L}_{\text{sdf}} + \lambda_{\text{free}} \cdot \mathcal{L}_{\text{free}}
\end{align}
}

\subsection{Outdoor Scenario}
\label{sec:method_outdoor_scenario}

To further show the generalization ability of our pre-training paradigm, we also have applied our methodology to the \revise{multi-modal outdoor autonomous driving scenario, where multi-view images and LiDAR point clouds are both available.}
To make the pre-training method suitable for these inputs, we convert them into the 3D volumetric space.

Specifically, for the LiDAR point clouds, we follow the same process in Sec.~\ref{sec:method_pipeline_overview} to augment the point clouds and voxelize the point features extracted by a 3D backbone.

For multi-view images $\sI = \{\mI_1, \mI_2, ...\}$, inspired by MAE~\cite{he2022masked}, we first mask out \revise{a ratio of patches} as the data augmentation to get $\hat{\sI}$. Then we leverage a 2D backbone $f_e^{(2d)}$ to extract multi-view image features $\tF_\mathrm{image} = f_e^{(2d)}(\hat{\sI})$.
The 2D features are subsequently transformed into the 3D ego-car coordinate system to obtain the 3D dense volume features.
Concretely, we first pre-define the 3D voxel coordinates $X_p\in\N^{l_x \times l_y \times l_z \times 3}$, and then project $X_p$ on multi-view images to index the corresponding 2D features.
The process can be calculated by:
\begin{equation}
    \tF_{\mathrm{dense}}=\tB(T_{\mathrm{c2i}} T_{\mathrm{l2c}} X_p, \tF_\mathrm{image}),
\end{equation}
where $T_\mathrm{l2c}$ and $T_{\mathrm{c2i}}$ denote the transformation matrices from the LiDAR coordinate system to the camera frame and from the camera frame to image coordinates, respectively, and $\tB$ represents the bilinear interpolation. The encoding process of the multi-view image case is shown in \figref{fig:overall_stru} (b).

\section{Experiments for Indoor Scenarios}
\revise{We summarize all our experimental settings in \tabref{tab:setting_summary}.} In this section, we first introduce comprehensive experiments on indoor datasets, which mainly contain two parts. In the first part, we use a lightweight backbone for ablation studies, which take as input multi-frame RGB-D only. We call this variant Ponder-RGBD. The subsequent part mainly focuses on a single, unified pre-trained model that pushes to the limits of performance, surpassing the previous SOTA pretraining pipeline substantially.

\begin{table*}[!tb]
\centering
\caption{\revise{Summary of different experimental settings.}}\label{tab:setting_summary}
\vspace{-0.8em}
\tablestyle{1pt}{1.05}
\begin{tabular}{lccccc}\toprule
&\multicolumn{2}{c}{Indoor} &\multicolumn{2}{c}{Outdoor} \\\cmidrule(lr){2-3}\cmidrule(lr){4-5}
&\secref{sec:rgbd} &\secref{sec:indoor} &\secref{sec:outdoor}& \\\cmidrule(lr){2-2}\cmidrule(lr){3-3}\cmidrule(lr){4-5}
Pre-train Datasets &ScanNet~\cite{dai2017scannet} &ScanNet~\cite{dai2017scannet}, S3DIS~\cite{armeni2017joint}, Structured3D~\cite{zheng2020structured3d} & NuScenes~\cite{caesar2020nuscenes}  & \\
Pre-train Data size &$\sim$1500 Scenes &$\sim$5271 Scenes &$\sim$700 scenes & \\
Total Pre-train Epochs &100 &200 &24 & \\
Total Pre-train Time &$\sim$4 days &$\sim$7 days &$\sim$2 days & \\
Input Format &Multi-View RGB-D Images &Scanned Scene Point Clouds &LiDAR Point Clouds / Multi-Camera Images & \\
Pre-trained Architecture &Encoder &Encoder &Encoder + BEV Volume & \\
Loss Functions &\eqref{loss:indoor_rgbd} &\eqref{loss:indoor} &\eqref{loss:render} & \\
\multirow{4}{*}{Downstream Tasks} & Object Detection, & Semantic Segmentation& Object Detection& \\
& Semantic Segmentation, & Instance Segmentation,& Semantic Segmentation& \\
& Scene Reconstruction, & Object Reconstruction,& \etc& \\
& Image Synthesis, \etc& \etc & & \\
\bottomrule
\end{tabular}
\end{table*}

\subsection{Indoor Scene Multi-frame RGB-D Images as Inputs}
\label{sec:rgbd}

\subsubsection{Experimental Setup}

We use ScanNet~\cite{dai2017scannet} RGB-D images as our pre-training dataset. ScanNet is a widely used real-world indoor dataset, which contains more than 1500 indoor scenes. Each scene is carefully scanned by an RGB-D camera, leading to about 2.5 million RGB-D frames in total. We follow the same \textit{train} / \textit{val} split with VoteNet~\cite{qi2019votenet}. In this part, we have not introduced semantic rendering yet, which will be used in the part of scene-level as inputs. \revise{We also provide a detailed explanation on how the multi-frame RGB-D images are lifted into point clouds in Appendix~\ref{appx:rgbd_lift}.}

\subsubsection{Implementation and Training Details}

For the version of RGB-D inputs, \minorrevise{the projected point cloud contains 3 input channels for RGB, with a grid size of} $\vg = 0.02^3$ \revisenew{meters}. After densifying $\tV$ to $64 \times 64 \times 64$ with $96$ channels, we apply a dense UNet with an output channel number of $128$. $\phi_\mathrm{SDF}$ is designed as a $5$-layer MLP while $\phi_\mathrm{RGB}$ is a $3$-layer MLP.

During pre-training, a mini-batch of batch size $8$ includes point clouds from $8$ scenes. Each point cloud input to our sparse backbone is back-projected from $5$ continuous RGB-D frames of a video from ScanNet's raw data with a frame interval of $20$. The $5$ frames are also used as the supervision of the network.We randomly down-sample the input point cloud to 20,000 points and follow the masking strategy as used in Mask Point~\cite{liu2022maskpoint}. 

We train the proposed pipeline for 100 epochs using an AdamW optimizer~\cite{losh2019adamw} with a weight decay of $0.05$. 
The learning rate is initialized as $1e^{-4}$ with an exponential schedule. 
For the rendering process, we randomly choose 128 rays for each image and sample 128 points for each ray.

\subsubsection{Comparison Experiments}

\begin{table*}[!tb]
\centering
\caption{
  \textbf{Indoor Ponder-RGBD 3D object detection} mAP@25 and mAP@50 {on ScanNet and SUN RGB-D with 
\textbf{VoteNet}~\cite{qi2019votenet} backbone.} 
  The DepthContrast~\cite{zhang_depth_contrast} and Point-BERT~\cite{yu2022pointbert} results are adopted from IAE~\cite{yan2022implicit} and MaskPoint~\cite{liu2022maskpoint}.
  \textbf{Ponder-RGBD} outperforms both VoteNet-based and 3DETR-based point cloud pre-training methods with fewer training epochs.
  }
  \label{tab:3d object detection}
\vspace{-0.8em}
\tablestyle{5.25pt}{1.05}
\begin{tabular}{lcccccccc}\toprule
\multirow{2}{*}{Method} &\multirow{2}{*}{Detection Model} &\multirow{2}{*}{Pre-training Type} &\multirow{2}{*}{Pre-training Epochs} &\multicolumn{2}{c}{ScanNet Val} &\multicolumn{2}{c}{SUN RGB-D Val} \\\cmidrule(lr){5-6}\cmidrule(lr){7-8}
& & & &mAP@50 &mAP@25 &mAP@50 &mAP@25 \\\cmidrule{1-8}
3DETR~\cite{misra2021-3detr} &3DETR &- &- &37.5 &62.7 &30.3 &58 \\
+ Point-BERT~\cite{yu2022pointbert} &3DETR &Masked Auto-Encoding &300 &38.3 &61.0 &- &- \\
+ MaskPoint~\cite{liu2022maskpoint} &3DETR &Masked Auto-Encoding &300 &40.6 &63.4 &- &- \\\cmidrule{1-8}
VoteNet~\cite{qi2019votenet} &VoteNet &- &- &\underline{33.5} &\underline{58.6} &\underline{32.9} &\underline{57.7} \\
+ STRL~\cite{huang2021spatio} &VoteNet &Contrast &100 &38.4 &59.5 &35.0&58.2 \\
+ RandomRooms~\cite{rao2021randomrooms} &VoteNet &Contrast &300 &36.2 &61.3 &35.4 &59.2 \\
+ PointContrast~\cite{xie2020pointcontrast} &VoteNet &Contrast &- &38.0&59.2 &34.8 &57.5 \\
+ PC-FractalDB~\cite{yamada2022point} &VoteNet &Contrast &- &38.3 &61.9 &33.9 &59.4 \\
+ DepthContrast~\cite{zhang_depth_contrast} &VoteNet &Contrast &1000 &39.1 &62.1 &35.4 &60.4 \\
+ IAE~\cite{yan2022implicit} &VoteNet &Masked Auto-Encoding &1000 &39.8 &61.5 &36.0&60.4 \\
\cellcolor[HTML]{efefef}\textbf{+ Ponder-RGBD (Ours)} &\cellcolor[HTML]{efefef}VoteNet &\cellcolor[HTML]{efefef}Rendering &\cellcolor[HTML]{efefef}100 &\cellcolor[HTML]{efefef}\textbf{41.0\up{7.5}} &\cellcolor[HTML]{efefef}\textbf{63.6\up{5.0}} &\cellcolor[HTML]{efefef}\textbf{36.6\up{3.7}} &\cellcolor[HTML]{efefef}\textbf{61.0\up{3.3}} \\
\bottomrule
\end{tabular}
\end{table*}

\begin{table}[!tb]
\centering
\caption{
  \textbf{Indoor Ponder-RGBD 3D object detection} mAP@25 and mAP@50 {on ScanNet validation set} with \textbf{H3DNet}~\cite{zhang2020h3dnet} backbone. Ponder-RGBD significantly boosts the accuracy by a margin of +2.8 and +1.2 for mAP@50 and mAP@25, respectively.
  }
  \label{tab:3d detection h3dnet}
\vspace{-0.8em}
\tablestyle{17pt}{1.05}
\begin{tabular}{lccc}\toprule
Method &mAP@50 &mAP@25 \\\cmidrule{1-3}
VoteNet~\cite{qi2019votenet} &33.5 &58.7 \\
3DETR~\cite{misra2021-3detr} &37.5 &62.7 \\
3DETR-m~\cite{misra2021-3detr} &47.0 &65.0 \\
\rowcolor[gray]{.96}
H3DNet~\cite{zhang2020h3dnet} &\underline{48.1} &\underline{67.2} \\
\rowcolor[gray]{.92}
\textbf{+ Ponder-RGBD (Ours)} &\textbf{50.9\up{2.8}} &\textbf{68.4\up{1.2}} \\
\bottomrule
\end{tabular}
\end{table}

\noindent \textbf{Object Detection}
We select two representative approaches, VoteNet~\cite{qi2019votenet} and H3DNet~\cite{zhang2020h3dnet}, as the baselines. VoteNet leverages a voting mechanism to obtain object centers, which are used for generating 3D bounding box proposals.
By introducing a set of geometric primitives, H3DNet achieves a significant improvement in accuracy compared to previous methods.
Two datasets are applied to verify the effectiveness of our method: ScanNet~\cite{dai2017scannet} and SUN RGB-D~\cite{song2015sun}.
Different from ScanNet, which contains fully reconstructed 3D scenes, 
SUN RGB-D is a single-view RGB-D dataset with 3D bounding box annotations. It has 10,335 RGB-D images for 37 object categories. 
For pre-training, we use PointNet++ as the point cloud encoder $f_e^{(s)}$, which is identical to the backbone used in VoteNet and H3DNet.
We pre-train $f_e^{(s)}$ on the ScanNet dataset and transfer the weight as the downstream initialization.  
Following~\cite{qi2019votenet}, we use average precision with 3D detection IoU threshold $0.25$ and threshold $0.5$ as the evaluation metrics.

The 3D detection results are shown in \tabref{tab:3d object detection}. Our method improves the baseline of VoteNet without pre-training by a large margin, boosting mAP@50 by 7.5\% and 3.7\% for ScanNet and SUN RGB-D, respectively.
IAE~\cite{yan2022implicit} is a pre-training method that represents the inherent 3D geometry in a continuous manner. 
Our learned point cloud representation achieves higher accuracy because it is able to recover both the geometry and appearance of the scene.  The mAP@50 and mAP@25 of our method are higher than that of IAE by 1.2\% and 2.1\% on ScanNet, respectively. 
Besides, we have observed that our method surpasses the recent point cloud pre-training approach, MaskPoint~\cite{liu2022maskpoint}, even when using a less sophisticated backbone (PointNet++ vs. 3DETR), as presented in \tabref{tab:3d object detection}. 
To verify the effectiveness of Ponder-RGBD, we also apply it for a much stronger baseline, H3DNet. As shown in \tabref{tab:3d detection h3dnet}, our method surpasses H3DNet by +2.8 and +1.2 for mAP@50 and mAP@25, respectively.

\noindent \textbf{Semantic Segmentation}
3D semantic segmentation is another fundamental scene understanding task. 
We select one of the top-performing backbones, MinkUNet~\cite{choy20194d}, for finetuning. MinkUNet leverages 3D sparse convolution to extract effective 3D scene features. We report the finetuning results on the ScanNet dataset with the mean IoU of the validation set as the evaluation metric. \tabref{tab:3D semantic segmentation mink} shows the quantitative results of Ponder-RGBD with MinkUNet. The results demonstrate that Ponder-RGBD is effective in improving the semantic segmentation performance, achieving a significant improvement of 1.3 mIoU.

\noindent \textbf{Scene Reconstruction}
3D scene reconstruction task aims to recover the scene geometry, e.g. mesh, from the point cloud input. 
We choose ConvONet~\cite{peng2020convolutional} as the baseline model, 
whose architecture is widely adopted in~\cite{chibane2020implicit, liu2020neural, yu2021plenoctrees}.
Following the same setting as ConvONet, we conduct experiments on the Synthetic Indoor Scene Dataset (SISD)~\cite{peng2020convolutional}, which is a synthetic dataset and contains 5000 scenes with multiple ShapeNet~\cite{chang2015shapenet} objects. 
To make a fair comparison with IAE~\cite{yan2022implicit}, we use the same VoteNet-style PointNet++ as the encoder of ConvONet, which down-samples the original point cloud to 1024 points. 
Following~\cite{peng2020convolutional}, we use Volumetric IoU, Normal Consistency (NC), and F-Score~\cite{tatarchenko2019single} with the threshold value of 1\% as the evaluation metrics.
The results are shown in \tabref{tab:3d scene reconstruction}. Compared to the baseline ConvONet model with PointNet++, IAE is not able to boost the reconstruction results, while the proposed approach can improve the reconstruction quality (+2.4\% for IoU). 
The results show the effectiveness of Ponder-RGBD for the 3D scene reconstruction task.

\begin{table}[!tb]\centering
\caption{
  \textbf{Indoor Ponder-RGBD 3D segmentation} mIoU on ScanNet validation dataset. 
  }
  \label{tab:3D semantic segmentation mink}
  \vspace{-0.8em}
\tablestyle{33pt}{1.05}
\begin{tabular}{lcc}\toprule
Method &Val mIoU \\\cmidrule{1-2}
PointNet++~\cite{qi2017pointnet++} &53.5 \\
KPConv~\cite{thomas2019KPConv} &69.2 \\
SparseConvNet~\cite{3DSemanticSegmentationWithSubmanifoldSparseConvNet} &69.3 \\
PT~\cite{zhao2021pointtransformer} &70.6 \\
\rowcolor[gray]{.96}
MinkUNet~\cite{choy20194d} &\underline{72.2} \\
\rowcolor[gray]{.92}
\textbf{+ Ponder-RGBD (Ours)} &\textbf{73.5\up{1.3}} \\
\bottomrule
\end{tabular}
\vspace{-1.2em}
\end{table}

\begin{table}[!tb]\centering
\caption{
  \textbf{Indoor Ponder-RGBD 3D scene reconstruction} \textit{IoU}, \textit{NC}, and \textit{F-Score} on SISD dataset with \textbf{PointNet++}. 
  }
  \label{tab:3d scene reconstruction}
\vspace{-0.8em}
\tablestyle{3.5pt}{1.05}
\begin{tabular}{lccccc}\toprule
Method &Encoder &IoU$\uparrow$ &NC$\uparrow$ &F-Score$\uparrow$ \\\cmidrule{1-5}
IAE~\cite{yan2022implicit} &PointNet++ &75.7 &88.7 &91.0 \\
\rowcolor[gray]{.96}
ConvONet~\cite{peng2020convolutional} &PointNet++ &\underline{77.8} &\underline{88.7} &\underline{90.6} \\
\rowcolor[gray]{.92}
\textbf{+ Ponder-RGBD (Ours)} &PointNet++ &\textbf{80.2\up{2.4}} &\textbf{89.3\up{0.6}} &\textbf{92.0\up{1.4}} \\
\bottomrule
\end{tabular}
\vspace{-1.2em}
\end{table}

\noindent \textbf{Image Synthesis From Point Clouds}
\revise{
We validate our method for image synthesis from point clouds using Point-NeRF~\cite{xv2022pointnerf}, focusing on the generalizable setting with the DTU dataset~\cite{jensen2014large}. By replacing 2D image features with point features from a DGCNN network, our approach improves PSNR and accelerates convergence. Details and visualizations are in Appendix~\ref{appx:rgb_img}.
}

\subsubsection{Ablation Study}

\noindent \textbf{Influence of Rendering Targets} 
The rendering part of our method contains two items: RGB color image and depth image. We study the influence of each item with the transferring task of 3D detection. The results are presented in \tabref{tab:ablation study supervision}. 
Combining depth and color images for reconstruction shows the best detection results. In addition, using depth reconstruction presents better performance than color reconstruction for 3D detection.

\noindent \textbf{Influence of Mask Ratio} 
To augment point cloud data, we employ random masking as one of the augmentation methods, which divides the input point cloud into 2048 groups with 64 points. 
In this ablation study, we evaluate the performance of our method with different mask ratios, ranging from 0\% to 90\%, on the ScanNet and SUN RGB-D datasets, and report the results in \tabref{tab:supp mask ratio}. Notably, we find that even when no dividing and masking strategy is applied (0\%), our method achieves a competitive mAP@50 performance of 40.7 and 37.3 on ScanNet and SUN RGB-D, respectively.
Our method achieves the best performance on ScanNet with a mask ratio of 75\% and a mAP@50 performance of 41.7.
Overall, these results suggest that our method is robust to the hyper-parameter of mask ratio and can still achieve competitive performance without any mask operation.

\noindent \textbf{Influence of 3D Feature Volume Resolution} 
\revise{By default, we use a relatively small resolution of 3D feature volume. To explore the impact of resolution, we perform ablation studies with varying resolutions, as shown in Table \ref{tab:supp resolution}.
Inspired by advances in 3D reconstruction, we implement multi-resolution of [16, 32, 64] to make the volume with a much larger resolution. Results indicate that using a smaller resolution of 16, while potentially affecting rendering quality, can enhance downstream performance, accelerate pre-training, and reduce memory consumption.
}

\begin{table}[!tb]\centering
\centering
    \caption{
  \textbf{Indoor Ponder-RGBD ablation study for supervision type}.
  $\textit{3D detection}$ $\textit{{AP}}_{50}$ {on ScanNet and SUN RGB-D validation set.} 
  Combining color supervision and depth supervision can lead to better detection performance than using a single type of supervision.
  }
  \label{tab:ablation study supervision}
  \vspace{-0.8em}
\tablestyle{8pt}{1.05}
\begin{tabular}{lccc}\toprule
Supervision &ScanNet mAP@50 &SUN RGB-D mAP@50 \\\cmidrule{1-3}
VoteNet &\underline{33.5} &\underline{32.9} \\
+ Depth &40.9\up{7.4} &36.1\up{3.2} \\
+ Color &40.5\up{7.0} &35.8\up{2.9} \\
\rowcolor[gray]{.92}
\textbf{+ Depth + Color} &\textbf{41.0\up{7.5}} &\textbf{36.6\up{3.7}} \\
\bottomrule
\end{tabular}
\end{table}

\begin{table}[!tb]\centering
\caption{
  \textbf{Ablation study for mask ratio.} 
  $\textit{3D detection}$ $\textit{{AP}}_{50}$ {on ScanNet and SUN RGB-D validation set.} 
  }
  \label{tab:supp mask ratio}
  \vspace{-0.8em}
\tablestyle{11.5pt}{1.05}
\begin{tabular}{lccc}
\toprule
Mask Ratio &ScanNet mAP@50 &SUN RGB-D mAP@50 \\\cmidrule{1-3}
VoteNet &\underline{33.5} &\underline{32.9} \\
0\% &40.7\up{7.2} &37.3\up{4.4} \\
25\% &40.7\up{7.2} &36.2\up{3.3} \\
50\% &40.3\up{6.8} &36.9\up{4.0} \\
\rowcolor[gray]{.92}
\textbf{75\%} &\textbf{41.7\up{8.2}} &\textbf{37.0\up{4.1}} \\
90\% &41.0\up{7.5} &36.6\up{3.7} \\
\bottomrule
\end{tabular}
\end{table}

\begin{table}[!tb]\centering
\caption{
  \textbf{Ablation study for feature volume resolution.} 
  $\textit{3D detection}$ $\textit{{AP}}_{50}$ {on ScanNet and SUN RGB-D validation set.} 
  }
  \label{tab:supp resolution}
  \vspace{-0.8em}
\tablestyle{11pt}{1.05}
\begin{tabular}{lccc}\toprule
Resolution &ScanNet mAP@50 &SUN RGB-D mAP@50 \\\cmidrule{1-3}
VoteNet &\underline{33.5} &\underline{32.9} \\
16 &40.7\up{7.2} &36.6\up{3.7} \\
\rowcolor[gray]{.92}
\textbf{16 + 32 + 64} &\textbf{41.0\up{7.5}} &\textbf{36.6\up{3.7}} \\
\bottomrule
\end{tabular}
\end{table}

\begin{table}[!tb]\centering
\caption{
  \textbf{Ablation study for view number.} 
  $\textit{3D detection}$ $\textit{{AP}}_{50}$ {on ScanNet and SUN RGB-D validation set.} 
  Using multi-view supervision for pre-training can achieve better performance.
  }
\label{tab:ablation study multiview}
\vspace{-0.8em}
\tablestyle{13pt}{1.05}
\begin{tabular}{lccc}\toprule
\#View &ScanNet mAP@50 &SUN RGB-D mAP@50 \\\cmidrule{1-3}
VoteNet &\underline{33.5} &\underline{32.9} \\
1 &40.1\up{6.6} &35.4\up{2.5} \\
3 &40.8\up{7.3} &36.0\up{3.1} \\
\rowcolor[gray]{.92}
\textbf{5} &\textbf{41.0\up{7.5}} &\textbf{36.6\up{3.7}} \\
\bottomrule
\end{tabular}
\end{table}

\noindent \textbf{\minorrevise{Number of Input RGB-D Views}.} 
Our method utilizes $N$ RGB-D images, where $N$ is the input view number. We study the influence of $N$ and conduct experiments on 3D detection, as shown in \tabref{tab:ablation study multiview}. We change the number of input views while keeping the scene number of a batch still 8.
Using multi-view supervision helps to reduce single-view ambiguity. Similar observations are also found in the multi-view reconstruction task~\cite{long2022sparseneus}. 
Compared with the single view, multiple views achieve higher accuracy, boosting mAP@50 by 0.9\% and 1.2\% for ScanNet and SUN RGB-D datasets, respectively.

\subsubsection{Other Applications}

\begin{figure}[t]
  \centering
  \includegraphics[width=1.0\linewidth,trim=0 0 0 0]{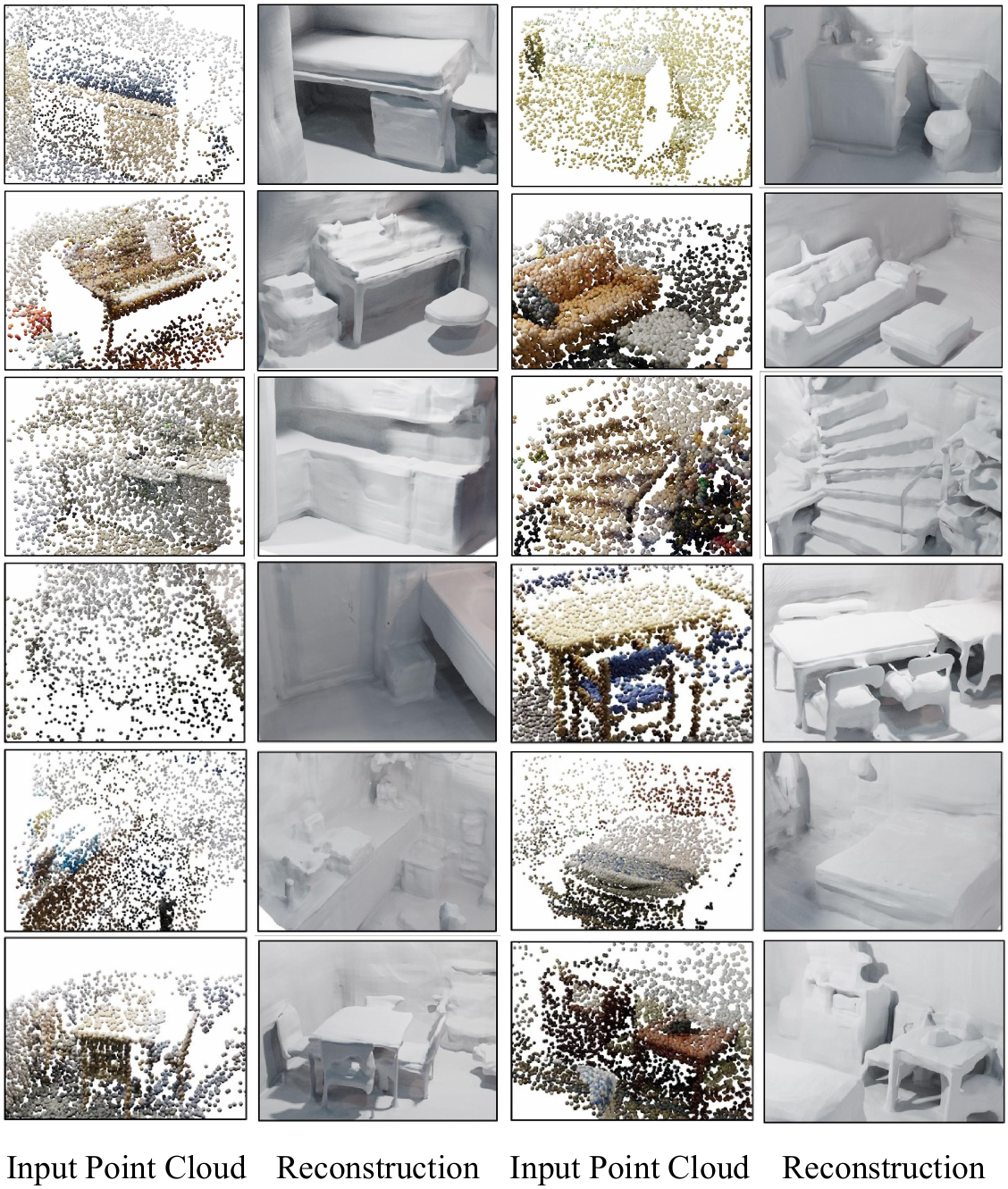}
  \vspace{-1.8em}
  \caption{\textbf{Reconstructed surface by Ponder-RGBD.} Our pre-training method can be easily integrated into the task of 3D reconstruction. 
  Despite the sparsity of the input point cloud (only 2\% points are used), our method can still recover precise geometric details.
}\label{fig:application reconstruction}
  \vspace{-0.3cm}
\end{figure}
The pre-trained model from our pipeline Ponder-RGBD itself can also be directly used for surface reconstruction from sparse point clouds. 
Specifically, after learning the neural scene representation, we query the SDF value in the 3D space and leverage the Marching Cubes~\cite{lorensen1987marching} to extract the surface.
We show the reconstruction results in \figref{fig:application reconstruction}. The results show that even though the input is sparse point clouds from complex scenes, our method is capable of recovering high-fidelity meshes.

\subsection{Indoor Scene Point Clouds as Inputs}
\label{sec:indoor}
\begin{table*}[!tb]
\centering
\caption{\textbf{Indoor semantic segmentation results.} Our method builds on SparseUNet~\cite{choy20194d}, and is
evaluated on ScanNet~\cite{dai2017scannet}, ScanNet200~\cite{rozenberszki2022language}, and S3DIS~\cite{armeni20163d} benchmarks. Compared to other pre-training approaches, \sexyname has significant finetuning improvements across all the benchmarks with shared pre-trained weights.}
\label{tab:indoor_sem_seg}
\vspace{-0.8em}
\tablestyle{11pt}{1.05}
\begin{tabular}{lccccccc}\toprule
\multirow{2}{*}{Method}&\multirow{2}{*}{\#Params.}&\multicolumn{2}{c}{ScanNet} &\multicolumn{2}{c}{ScanNet200} &\multicolumn{2}{c}{S3DIS} \\
\cmidrule(lr){3-4} \cmidrule(lr){5-6} \cmidrule(lr){7-8}
 &&Val mIoU&Test mIoU &Val mIoU &Test mIoU &Area5 mIoU &6-fold mIoU \\\cmidrule{1-8}
StratifiedFormer~\cite{lai2022stratified} & 18.8M&74.3 &73.7 &- &- &72.0 &78.1 \\
PointNeXt~\cite{qian2022pointnext} & 41.6M&71.5 &71.2 &- &- &70.5 &77.2 \\
PTv1~\cite{zhao2021pointtransformer} &11.4M &70.6&-  &27.8 &- &70.4 &76.5 \\
PTv2~\cite{wu2022pointtransv2} &12.8M &75.4 &75.2 &30.2 &- &71.6 &77.9 \\ \cmidrule{1-8}
SparseUNet~\cite{choy20194d} & 39.2M&\underline{72.2}&\underline{73.6} &\underline{25} &\underline{25.3} &\underline{65.4} &\underline{65.4} \\
+ PC~\cite{xie2020pointcontrast} &39.2M &74.1 &-&26.2 &- &70.3 &76.9 \\
+ CSC~\cite{hou2021exploring} &39.2M &73.8 &-&26.4 &24.9 &72.2 &- \\
+ MSC~\cite{wu2023masked} &39.2M &75.5&- &28.8 &- &70.1 &77.2 \\
\rowcolor[gray]{.96}
+ PPT~\cite{wu2023ppt} &41.0M &76.4 &76.6&31.9 &33.2 &72.7 &78.1 \\
\rowcolor[gray]{.92}
+ \textbf{\sexyname (Ours)} &41.0M &\textbf{77.0\up{4.8}}&\textbf{78.5\up{4.9}} &\textbf{32.3\up{7.3}} &\textbf{34.6\up{9.3}} &\textbf{73.2\up{7.8}} &\textbf{79.9\up{14.5}} \\
\bottomrule
\end{tabular}
\vspace{-1.2em}
\end{table*}

\subsubsection{Experimental Setup}
\revise{
In this setting, we aim to pre-train a unified backbone for various downstream tasks using whole scene point clouds, ensuring a consistent input and encoder stage. We use SparseUNet~\cite{choy20194d}, based on MinkUNet~\cite{choy20194d} optimized by SpConv~\cite{spconv2022}, due to its efficiency.
We focus on three main datasets: ScanNet~\cite{dai2017scannet}, S3DIS~\cite{armeni2017joint} and Structured3D~\cite{zheng2020structured3d}. We incorporate Point Prompt Training~\cite{wu2023ppt} (PPT), which assigns a batch norm layer to each dataset, as our baseline. 
After pre-training, we discard the rendering decoder, using the encoder's weights for downstream tasks, with or without task-specific heads. Please refer to Appendix~\ref{appx:indoor_setups} for more setup details. 
}

\subsubsection{Implementation and Training Details}
For the version of the unified backbone, we base our indoor experiments on Pointcept~\cite{pointcept2023}, a powerful and flexible codebase for point cloud perception research.  \revise{All hyper-parameters are the same as scratched PPT for fair comparison. The number of input channels for this setting is $6$, containing $3$ channels of colors and $3$ channels of surface normals. Grid size $\vg$ is $0.02^3$ meters. We apply common transformations including random dropout (at a mask ratio of $0.8$), rotation, scaling, and flipping.}

\revise{To improve pre-training efficiency and boost downstream performance, despite a reduction in rendering quality, we use a lightweight decoder. Specifically, we use a tiny dense 3D UNet with channels of $128$ after densifying. The dense feature volume $\tV$ is configured to be $128 \times 128 \times 32$ with $128$ channels. $\phi_{\mathrm{SDF}}$ is a shallow MLP with $3$ layers, and both $\phi_{\mathrm{RGB}}$ and $\phi_{\mathrm{SEMANTIC}}$ are $1$-layer MLPs. All shallow MLPs have a hidden channel of $128$. For semantic supervision, we use the text encoder of a "ViT-B/16" CLIP model, whose output semantic features have $512$ channels.}

\revise{In each scene-level input point cloud, we sample $5$ RGB-D frames for supervision, from which we sample $128$ rays each. Thus, a total of $128 \times 5 = 640$ pixel values per point cloud are supervised in each iteration. The weight coefficient $\lambda_C$ is $1.0$, and $\lambda_D$ is $0.1$.}

\revise{We train for $200$ epochs using an SGD optimizer with a weight decay of $1e^{-4}$ and a momentum of $0.9$. The learning rate is initialized at $8e^{-4}$ with a one-cycle scheduler~\cite{smith2019super}. For the rendering process, we randomly choose $5$ frames for each point cloud, $128$ rays for each image, and sample $128$ points for each ray. The batch size of point clouds is $64$. Models are trained on $8$ NVIDIA A100 GPUs.}

\subsubsection{Comparison Experiments}

\begin{table*}[!tp]\centering
\caption{\textbf{Indoor S3DIS semantic segmentation 6-fold cross-validation results.} \sexyname achieves the best average performance on all metrics including mIoU, mAcc and allAcc.}\label{tab:s3dis_six_fold}
\vspace{-0.8em}
\tablestyle{3.6pt}{1.05}
\begin{tabular}{lcccccccccccccccc}\toprule
\multirow{2}{*}{Metric} &\multicolumn{2}{c}{Area1} &\multicolumn{2}{c}{Area2} &\multicolumn{2}{c}{Area3} &\multicolumn{2}{c}{Area4} &\multicolumn{2}{c}{Area5} &\multicolumn{2}{c}{Area6} &\multicolumn{3}{c}{Average} \\\cmidrule(lr){2-3} \cmidrule(lr){4-5} \cmidrule(lr){6-7}\cmidrule(lr){8-9} \cmidrule(lr){10-11} \cmidrule(lr){12-13}\cmidrule(lr){14-16}%
&PPT &\cellcolor[HTML]{efefef}\sexyname &PPT &\cellcolor[HTML]{efefef}\sexyname &PPT &\cellcolor[HTML]{efefef}\sexyname &PPT &\cellcolor[HTML]{efefef}\sexyname &PPT &\cellcolor[HTML]{efefef}\sexyname &PPT &\cellcolor[HTML]{efefef}\sexyname &PPT &\cellcolor[HTML]{efefef}\sexyname &Scratch \\\cmidrule{1-16}
mIoU &83.0 &84.1 &65.4 &71.7 &87.1 &85.1 &74.1 &73.4 &72.7 &73.2 &86.4 &87.4 &78.1 &\textbf{79.9\up{1.8}} &65.4 \\
mAcc &90.3 &90.8 &75.6 &79.9 &91.8 &90.9 &84.0 &81.9 &78.2 &79.0 &92.5 &92.8 &85.4 &\textbf{86.5\up{1.1}} &- \\
allAcc &93.5 &93.7 &88.3 &90.1 &94.6 &94.2 &90.8 &90.8 &91.5 &92.2 &94.5 &94.8 &92.2 &\textbf{92.5}\up{0.3} &- \\
\bottomrule
\end{tabular}
\end{table*}
\noindent \textbf{Semantic Segmentation} 
We conduct indoor semantic segmentation experiments on ScanNet, S3DIS and Structured3D. We take PPT, the SOTA method using SparseUNet~\cite{choy20194d}, as our baseline, and report the comparison performance of scratch and finetuned models in \tabref{tab:indoor_sem_seg}. The S3DIS dataset contains 6 areas, among which we usually take Area 5 as the validation set. It is also common to evaluate 6-fold performance on it, so we also report the detailed 6-fold results in \tabref{tab:s3dis_six_fold}. Note that to avoid information leaks, we pre-train a new model on only ScanNet and Structured3D before finetuning on each area of the S3DIS dataset. The semantic segmentation experiments all show the significant performance of our proposed paradigm.

\noindent \textbf{Ablation on Semantic Supervision} Our pre-training typically employs ground-truth labels for semantic rendering. However, we demonstrate that PonderV2 achieves excellent performance even in a fully unsupervised setting. During pre-training, we use the semantic feature map from the visual encoder of CLIP (ViT-B/16) for supervision. This approach yields an mAP of $76.95$ on ScanNet, closely matching the $77.0$ mAP achieved with ground-truth labels and surpassing all other baselines.

\noindent \textbf{Instance Segmentation} %
3D instance segmentation is another fundamental perception task. We benchmark our fine-tuning resultsfor 3D instance segmentation on ScanNet, ScanNet200, and S3DIS (Area 5), as shown in \tabref{tab:indoor_insseg}. The results show that our pre-training approach also helps enhance instance segmentation understanding.

\noindent \textbf{Data Efficiency on ScanNet}
The ScanNet benchmark also contains data efficiency settings of limited annotations (LA) and limited reconstructions (LR). 
In the LA setting, the models are allowed to see only a small ratio of labels while in the LR setting, the models can only see a small number of reconstructions (scenes). Again, to prevent information leaking, we pre-train our model on only S3DIS and Structured3D before finetuning. Results in \tabref{tab:indoor_data_eff} indicate that our approach is more data efficient compared to the baseline.

\begin{table*}[!tb]\centering
\caption{\textbf{Indoor data efficient results.} We follow the ScanNet Data Efficient benchmark~\cite{hou2021exploring} and compare the validation results of \sexyname with previous pre-training methods. All methods are trained with SparseUNet. \textit{Pct.} in \textbf{limited reconstructions} setting denotes the percentage of scene reconstruction that could be used for training. \textit{\#Pts.} in \textbf{limited annotations} setting denotes the number of points per scene that are annotated for training.}
\label{tab:indoor_data_eff}
\vspace{-0.8em}
\tablestyle{8.5pt}{1.05}
\begin{tabular}{cccccc|cccccc}\toprule
\multirow{2}{*}{Pct.}&\multicolumn{5}{c|}{Limited Reconstructions} & \multirow{2}{*}{\#Pts.}&\multicolumn{5}{c}{Limited Annotations} \\\cmidrule(lr){2-6}\cmidrule(lr){8-12}
 &Sratch & CSC~\cite{hou2021exploring}& MSC~\cite{wu2023masked}&PPT &\cellcolor[HTML]{efefef}\textbf{\sexyname} & &Sratch &CSC~\cite{hou2021exploring} & MSC~\cite{wu2023masked}&PPT &\cellcolor[HTML]{efefef}\textbf{\sexyname} \\\cmidrule{1-12}
1\% &26.0 & 28.9&29.2 &31.3 &\textbf{34.6\up{8.6}} &20 &41.9 &55.5 & 60.1&60.6 &\textbf{67.0\up{25.1}} \\
5\% &47.8 & 49.8&59.4 &52.2 &\textbf{56.5\up{8.7}} &50 &53.9 &60.5 &66.8 &67.5 &\textbf{72.2\up{18.3}} \\
10\% &56.7 &59.4 & 61.0&62.8 &\textbf{66.0\up{9.3}} &100 &62.2 &65.9 &69.7 &70.8 &\textbf{74.3\up{12.1}} \\
20\% &62.9 & 64.6&64.9 &66.4 &\textbf{68.2\up{5.3}} &200 &65.5 &68.2 &70.7 &72.2 &\textbf{74.8\up{9.3}} \\
100\% &72.2 & 73.8&75.3 &76.4 &\textbf{77.0\up{4.8}} &Full &72.2 &73.8 &75.3 &76.4 &\textbf{77.0}\up{4.8} \\
\bottomrule
\end{tabular}
\end{table*}

\begin{table}[!tb]
\centering
\caption{\revise{\textbf{Indoor instance segmentation results.} We fine-tune \sexyname on ScanNet and ScanNet200 using PointGroup~\cite{jiang2020pointgroup}. Our method outperforms previous pre-training methods in mAP@25, mAP@50, and overall mAP.}}\label{tab:indoor_insseg}
\vspace{-0.8em}
\resizebox{\linewidth}{!}{
\tablestyle{1.5pt}{1.05}
\begin{tabular}{lcccccc}\toprule
\multirow{2}{*}{Method}&\multicolumn{3}{c}{Scannet } &\multicolumn{3}{c}{ScanNet200} \\
\cmidrule(lr){2-4}\cmidrule(lr){5-7}
 &mAP@25 &mAP@50 &mAP &mAP@25 &mAP@50 &mAP  \\\cmidrule{1-7}
PointGroup~\cite{jiang2020pointgroup} &\underline{72.8} &\underline{56.9} &\underline{36.0} &\underline{32.2} &\underline{24.5} &\underline{15.8}  \\
+ PC~\cite{xie2020pointcontrast} &- &58.0 &- &- &24.9 &-  \\
+ CSC~\cite{hou2021exploring} &- &59.4 &- &- &25.2 &-  \\
+ LGround~\cite{rozenberszki2022language} &- &- &- &- &26.1 &-  \\
+ MSC~\cite{wu2023masked} &74.7 &59.6 &39.3 &34.3 &26.8 &17.3  \\
\rowcolor[gray]{.96}
+ PPT~\cite{wu2023ppt} &76.9 &62.0 &40.7 &36.8 &29.4 &19.4  \\
\rowcolor[gray]{.92}
 + \textbf{\sexyname (Ours)} &\textbf{77.0\up{4.2}} &\textbf{62.6\up{5.7}} &\textbf{40.9\up{4.9}} &\textbf{37.6\up{5.4}} &\textbf{30.5\up{6.0}} &\textbf{20.1\up{4.3}} \\
\bottomrule
\end{tabular}
}
\end{table}

\noindent \textbf{Object Reconstruction} 
\revise{We tested \sexyname for object reconstruction using MCC~\cite{wu2023multiview} on CO3D~\cite{reizenstein2021common}. Details and results can be found in Appendix~\ref{appx:indoor_rec}, showing the effective transfer of scene-level knowledge to object-level tasks.}

\section{Experiments for Outdoor Scenarios}
\label{sec:outdoor}
\subsection{Experimental Setup}
\revise{
In this section, we detail \sexyname's outdoor experiments.
We conducted experiments on the challenging NuScenes~\cite{caesar2020nuscenes} dataset for outdoor autonomous driving, focusing on 3D object detection and semantic segmentation tasks.}
It consists of 700 scenes for training, 150 scenes for validation, and 150 scenes for testing.
Each scene is captured through six different cameras, providing images with surrounding views, and is accompanied by a point cloud from LiDAR.
The dataset comes with diverse annotations, supporting tasks like 3D object detection and 3D semantic segmentation. 
For detection evaluation, we employ nuScenes detection score (NDS) and mean average precision (mAP), and for segmentation assessment, we use mean intersection-over-union (mIoU).

\subsection{Implementation and Training Details}
We base our code on the MMDetection3D~\cite{mmdet3d2020} toolkit and train all models on 4 NVIDIA A100 GPUs. The input image is configured to $1600 \times 900$ pixels, while the voxel dimensions $\vg$ for point cloud voxelization are $[0.075, 0.075, 0.2]$.
During the pre-training phase, we implemented several data augmentation strategies, such as random scaling and rotation. Additionally, we partially mask the inputs, focusing only on visible regions for feature extraction. The masking size and ratio for images are configured to $32$ and $0.3$, and for points to $8$ and $0.8$, respectively.
For the segmentation task, we use SparseUNet as our sparse backbone $f_e^{(e)}$, and for the detection task, we use VoxelNet~\cite{yan2018second}, which is similar to the encoder part of SparseUNet, as our backbone.
For multi-image setting, we use ConvNeXt-small~\cite{liu2022convnext} as our feature extractor $f_e^{(2d)}$.
A uniform voxel representation $\tV$ with the shape of $180 \times 180 \times 5$ is constructed. The $f_d^{(d)}$ here is a $3$-kernel size convolution which serves as a feature projection layer reducing the $\tV$'s feature dimension to $32$. For the rendering decoders, we utilize a $6$-layer MLP for $\phi_\mathrm{SDF}$ and a $4$-layer MLP for $\phi_\mathrm{RGB}$. In the rendering phase, $512$ rays per image view and $96$ points per ray are randomly selected. We maintain the loss scale factors for $\lambda_\mathrm{RGB}$ and $\lambda_\mathrm{depth}$ at $10$. The model undergoes training for $12$ epochs using the AdamW optimizer with initial learning rates of $2e^{-5}$ and $1e^{-4}$ for point and image modalities, respectively.
In the ablation studies, unless explicitly stated, fine-tuning is conducted for $12$ epochs on 50\% of the image data and for $20$ epochs on 20\% of the point data, without the implementation of the CBGS~\cite{zhu2019cbgs} strategy.

\subsection{Outdoor Comparison Experiments}
\noindent \textbf{3D Object Detection}
In Table~\ref{tab:nuscene_val}, we compare \sexyname with previous detection approaches on the nuScenes validation set.
We adopt UVTR~\cite{li2022uvtr} as our baselines for point-modality (UVTR-L), camera-modality (UVTR-C), Camera-Sweep-modality(UVTR-CS) and fusion-modality (UVTR-M).
Benefits from the effective pre-training, \sexyname consistently improves the baselines, namely, UVTR-L, UVTR-C, and UVTR-M, by 2.9, 2.4, and 3.0 NDS, respectively.
When taking multi-frame cameras as inputs, \sexyname-CS brings 1.4 NDS and 3.6 mAP gains over UVTR-CS.
Our pre-training technique also achieves 1.7 NDS and 2.1 mAP improvements over the monocular-based baseline FCOS3D~\cite{wang2021fcos3d}.
Without any test time augmentation or model ensemble, our single-modal and multi-modal methods, \sexyname-L, \sexyname-C, and \sexyname-M, achieve impressive NDS of 70.6, 47.4, and 73.2, respectively, surpassing existing state-of-the-art methods.

\noindent \textbf{3D Semantic Segmentation}
In Table~\ref{tab:semseg}, we compare PonderV2 with previous point clouds semantic segmentation approaches on the nuScenes Lidar-Seg dataset. Benefits from the effective pre-training, PonderV2 improves the baselines by 6.1 mIoU, achieving state-of-the-art performance on the validation set. Meanwhile, PonderV2 achieves an impressive
mIoU of 81.1 on the test set, which is comparable with existing state-of-the-art methods.

\begin{table}[t]
	\centering
        \setlength{\tabcolsep}{4pt}
	\caption{
		\revise{Comparisons on the nuScenes {\em val} set {\em without} ensemble and test-time augmentation.
		$\dagger$: denotes our reproduced results based on  MMDetection3D~\cite{mmdet3d2020}.
		L, C, CS, and M indicate the LiDAR, Camera, Camera Sweep, and Multi-modality input, respectively.}
	}
	\vspace{-0.8em}
        \tablestyle{2pt}{1.05}
        \begin{tabular}{lcccccc}
		\toprule
		Methods & \revisenew{Presented} at & Modality & CS & CBGS & NDS$\uparrow$ & mAP$\uparrow$ \\
		\midrule
		PVT-SSD~\cite{yang2023pvtssd} & CVPR'23 & L & & \checkmark & 65.0 & 53.6 \\
		CenterPoint~\cite{yin2021center} & CVPR'21 & L & & \checkmark & 66.8 & 59.6 \\
		FSDv1~\cite{fan2022fsd} & NeurIPS'22 & L & &  \checkmark & 68.7 & 62.5 \\
		VoxelNeXt~\cite{chen2023voxelnext} & CVPR'23 & L & & \checkmark & 68.7 & 63.5 \\
		LargeKernel3D~\cite{chen2023largekernel3d} & CVPR'23 & L & & \checkmark & 69.1 & 63.3 \\
		TransFusion-L~\cite{bai2022transfusion} & CVPR'22 & L & & \checkmark & 70.1 & 65.1 \\
            CMT-L~\cite{yan2023cmt} & ICCV'23 & L & & \checkmark & 68.6 & 62.1 \\
		\rowcolor[gray]{.96}
		UVTR-L~\cite{li2022uvtr} & NeurIPS'22 & L & & \checkmark & 67.7 & 60.9 \\
		\rowcolor[gray]{.92}
		\textbf{+ \sexyname (Ours)} & - & L & & \checkmark & \textbf{70.6} & \textbf{65.0} \\
		\midrule
		BEVFormer-S~\cite{li2022bevformer} & ECCV'22 & C & & \checkmark & 44.8 & 37.5 \\
		SpatialDETR~\cite{doll2022spatialdetr} & ECCV'22 & C & & & 42.5 & 35.1 \\
		PETR~\cite{liu2022petr} & ECCV'22 & C & & \checkmark & 44.2 & 37.0 \\
		Ego3RT~\cite{lu2022ego3rt} & ECCV'22 & C & & & 45.0 & 37.5 \\
		3DPPE~\cite{shu20233dppe} & ICCV'23 & C & & \checkmark & 45.8 & 39.1 \\
		CMT-C~\cite{yan2023cmt} & ICCV'23 & C & & \checkmark & 46.0 & 40.6 \\
		\rowcolor[gray]{.96}
		FCOS3D$^\dag$~\cite{wang2021fcos3d} & ICCVW'21 & C & & & 38.4 & 31.1 \\
		\rowcolor[gray]{.92}
		\textbf{+ \sexyname (Ours)} & - & C & & & \textbf{40.1} & \textbf{33.2} \\
		\rowcolor[gray]{.96}
		UVTR-C~\cite{li2022uvtr} & NeurIPS'22  & C & & & 45.0 & 37.2 \\
		\rowcolor[gray]{.92}
		\textbf{+ \sexyname (Ours)} & - & C & & & \textbf{47.4} & \textbf{41.5} \\
		\rowcolor[gray]{.96}
		UVTR-CS~\cite{li2022uvtr} & NeurIPS'22 & C & \checkmark & & 48.8 & 39.2 \\
		\rowcolor[gray]{.92}
		\textbf{+ \sexyname (Ours)} & - & C & \checkmark & & \textbf{50.2} & \textbf{42.8} \\
		\midrule
		FUTR3D~\cite{chen2022futr3d} & arXiv'22 & C+L & & \checkmark & 68.3 & 64.5 \\
		PointPainting~\cite{vora2020pointpainting} & CVPR'20 & C+L & & \checkmark & 69.6 & 65.8 \\
		MVP~\cite{yin2021mvp} & NeurIPS'21 & C+L & & \checkmark & 70.8 & 67.1 \\
		TransFusion~\cite{bai2022transfusion} & CVPR'22 & C+L & & \checkmark & 71.3 & 67.5\\
		AutoAlignV2~\cite{chen2022autoalignv2} & ECCV'22 & C+L & & \checkmark & 71.2 & 67.1 \\
		BEVFusion~\cite{liang2022bevfusion} & NeurIPS'22 & C+L & & \checkmark & 71.0 & 67.9 \\
		BEVFusion~\cite{liu2023bevfusion} & ICRA'23 & C+L & & \checkmark & 71.4 & 68.5 \\
		DeepInteraction~\cite{yang2022deepinteraction} & NeurIPS'22 & C+L & & \checkmark & 72.6 & 69.9 \\
		CMT-M~\cite{yan2023cmt} & ICCV'23 & C+L & & \checkmark & 72.9 & 70.3 \\
		\rowcolor[gray]{.96}
		UVTR-M~\cite{li2022uvtr} & NeurIPS'22 & C+L & & \checkmark & 70.2 & 65.4 \\
		\rowcolor[gray]{.92}
		\textbf{+ \sexyname (Ours)} & - & C+L & & \checkmark & \textbf{73.2} & \textbf{69.9} \\
		\bottomrule
	\end{tabular}%
	\label{tab:nuscene_val}
\end{table}

\begin{table}[!t]
\centering
\caption{Comparisons of different methods with a single model on the nuScenes segmentation dataset.}
\vspace{-0.8em}
\tablestyle{8pt}{1.05}
\begin{tabular}{lccc}
    \toprule
     Methods & Backbone & Val mIoU & Test mIoU \\
     \midrule
     SPVNAS~\cite{tang2020spvnas} & Sparse CNN & - & 77.4 \\
     Cylinder3D~\cite{zhu2021cylindrical} & Sparse CNN & 76.1 & 77.2 \\
     SphereFormer~\cite{lai2023sphereformer} & Transformer & 78.4 & 81.9 \\
     \rowcolor[gray]{.96}
     SparseUNet~\cite{choy20194d} & Sparse CNN & 73.3 & - \\
     \rowcolor[gray]{.92}
     \textbf{+ \sexyname (Ours)} & Sparse CNN & \textbf{79.4} & \textbf{81.1} \\
    \bottomrule
\end{tabular}%
\label{tab:semseg}
\vspace{-0.2em}
\end{table}

\begin{table}[!t]
	\centering
	\caption{Comparison with camera-based pre-training.}
	\vspace{-0.8em}
        \tablestyle{10pt}{1.05}
        \begin{tabular}{lcccc}
		\toprule
		\multirow{2}{*}{Methods} & \multicolumn{2}{c}{Label} & \multirow{2}{*}{NDS} & \multirow{2}{*}{mAP} \\\cmidrule{2-3}
			& 2D & 3D & & \\
		\midrule
			UVTR-C (Baseline)  & & & \underline{25.2} & \underline{23.0} \\
			+ Depth Estimator & & & 26.9\up{1.7} & 25.1\up{2.1} \\
			+ Detector & \checkmark & & 29.4\up{4.2} & 27.7\up{4.7} \\
			+ 3D Detector & & \checkmark & 31.7\up{6.5} & 29.0\up{6.0} \\
            \rowcolor[gray]{.92}
		\textbf{+ \sexyname} & & & \textbf{32.9\up{7.7}} & \textbf{32.6\up{9.6}} \\
		\bottomrule 
	\end{tabular}%
	\label{tab:image_pre_train}
 \vspace{-0.2em}
\end{table}

\begin{table}[!t]
	\centering
	\caption{Comparison with point-based pre-training.}
	\vspace{-0.8em}
        \tablestyle{10.5pt}{1.05}
        \begin{tabular}{lcccc}
		\toprule
		\multirow{2}{*}{Methods} & \multicolumn{2}{c}{Support} & \multirow{2}{*}{NDS} & \multirow{2}{*}{mAP} \\\cmidrule(lr){2-3}
			& 2D & 3D & & \\
		\midrule
		UVTR-L (Baseline) & & & \underline{46.7} & \underline{39.0} \\
		+ Occupancy-based & & \checkmark & 48.2\up{1.5} & 41.2\up{2.2} \\
		+ MAE-based & & \checkmark & 48.8\up{2.1} & 42.6\up{3.6} \\
		+ Contrast-based & \checkmark & \checkmark & 49.2\up{2.5} & 48.8\up{9.8} \\
            \rowcolor[gray]{.92}
		\textbf{+ \sexyname} & \checkmark & \checkmark & \textbf{55.8\up{9.1}} & \textbf{48.1\up{9.1}} \\
		\bottomrule 
	\end{tabular}%
	\label{tab:point_pre_train}
 \vspace{-0.2em}
\end{table}

\noindent \textbf{Comparisons with Pre-training Methods}
\revise{
We conduct comparisons with camera-based pre-training approaches as well as point modality self-supervised methods in Table~\ref{tab:image_pre_train} and Table~\ref{tab:point_pre_train}. \sexyname outperforms several camera-based pre-training methods and achieves the best NDS performance in point modality comparisons. While it has slightly lower mAP than contrastive point-based methods, it simplifies the process by avoiding complex sample assignments. More details and analysis can be found in Appendix~\ref{appx:outdoor_pre}.
}

\subsection{Effectiveness on Various Backbones}
\revise{
Unlike most previous pre-training methods, our framework can be seamlessly applied to various modalities.
To verify the effectiveness of our approach, we set UVTR as our baseline, which contains detectors with point, camera, and fusion modalities.
Table~\ref{tab:diff_modality} shows the impact of \sexyname on different modalities.
\sexyname consistently improves the UVTR-L, UVTR-C, and UVTR-M by 9.1, 7.7, and 6.9 NDS, respectively.
We have also investigated that our approach is effective with different view transformation strategies and different backbone scales, as detailed in Appendix~\ref{appx:outdoor_eff}.
}

\begin{table}[!t]
	\centering
	\caption{Effectiveness on different modalities.}
	\vspace{-0.8em}
        \tablestyle{12pt}{1.05}
        \begin{tabular}{lccc}
		\toprule
		Methods & Modality & NDS & mAP \\
		\midrule
		UVTR-L & LiDAR & 46.7 & 39.0 \\
            \rowcolor[gray]{.92}
		\textbf{+ \sexyname} & LiDAR & \textbf{55.8\up{9.1}} & \textbf{48.1\up{9.1}} \\
		\midrule
            UVTR-C & Camera & 25.2 & 23.0 \\
            \rowcolor[gray]{.92}
		\textbf{+ \sexyname} & Camera & \textbf{32.9\up{7.7}} & \textbf{32.6\up{9.6}} \\
		\midrule
		UVTR-M & LiDAR-Camera & 49.9 & 52.7 \\
            \rowcolor[gray]{.92}
		\textbf{+ \sexyname} & LiDAR-Camera & \textbf{56.8\up{6.9}} & \textbf{57.0\up{4.3}} \\
		\bottomrule 
	\end{tabular}%
	\label{tab:diff_modality}
 \vspace{-1.2em}
\end{table}

\subsection{Ablation Studies}
\noindent \textbf{Masking Ratio}
Table~\ref{tab:mask_ratio} shows the influence of the masking ratio for the camera modality.
We discover that a masking ratio of 0.3, which is lower than the ratios used in previous MAE-based methods, is optimal for our method.
This discrepancy could be attributed to the challenge of rendering the original image from the volume representation, which is more complex compared to image-to-image reconstruction.
For the point modality, we adopt a mask ratio of 0.8, as suggested in \cite{yang2023gd-mae}, considering the spatial redundancy inherent in point clouds.

\noindent \textbf{Rendering Design}
Our examinations in Tables \ref{tab:decoder_depth}, \ref{tab:decoder_width}, and \ref{tab:rendering_tech} illustrate the flexible design of our differentiable rendering.
In Table~\ref{tab:decoder_depth}, we vary the depth $(D_\mathrm{SDF}, D_\mathrm{RGB})$ of the SDF and RGB decoders, revealing the importance of sufficient decoder depth for succeeding in downstream detection tasks.
This is because deeper ones may have the ability to adequately integrate geometry or appearance cues during pre-training.
Conversely, as reflected in Table~\ref{tab:decoder_width}, the width of the decoder has a relatively minimal impact on performance.
Thus, the default dimension is set to $32$ for efficiency.
Additionally, we explore the effect of various rendering techniques in Table~\ref{tab:rendering_tech}, which employ different ways for ray point sampling and accumulation.
Using NeuS~\cite{wang2021neus} for rendering records a 0.4 and 0.1 NDS improvement compared to UniSurf~\cite{oechsle2021unisurf} and VolSDF~\cite{lior2021volsdf} respectively, showcasing the learned representation can be improved by utilizing well-designed rendering methods and benefiting from the advancements in neural rendering.

\begin{table}[!t]
	\begin{minipage}[b]{0.48\linewidth}
	\centering
	\caption{Ablation study of the mask ratio.}
        \vspace{-0.8em}
        \tablestyle{8pt}{1.05}
        \begin{tabular}{x{20}x{24}x{24}}
        \toprule
        ratio & NDS & mAP \\
        \midrule
        0.1 & 31.9 & 32.4 \\
        \rowcolor[gray]{.92}
        \textbf{0.3} & \textbf{32.9} & \textbf{32.6} \\
        0.5 & 32.3 & \textbf{32.6} \\
        0.7 & 32.1 & 32.4 \\
        \bottomrule
        \end{tabular}%
	\label{tab:mask_ratio}
	\end{minipage}
	\hfill
	\begin{minipage}[b]{0.48\linewidth}
	\centering
	\caption{Ablation study of the decoder depth.}
        \vspace{-0.8em}
        \tablestyle{7pt}{1.05}
        \begin{tabular}{x{26}x{24}x{24}}\toprule
        layers & NDS & mAP \\
        \midrule
        (2, 2) & 31.3 & 31.3 \\
        (4, 3) & 31.9 & 31.6 \\
        (5, 4) & 32.1 & \cellcolor[HTML]{efefef}\textbf{32.7} \\
        (6, 4) & \cellcolor[HTML]{efefef}\textbf{32.9} & 32.6 \\\bottomrule
        \end{tabular}%
	\label{tab:decoder_depth}
	\end{minipage}
\end{table}

\begin{table}[!t]
	\begin{minipage}[b]{0.48\linewidth}
	\centering
	\caption{Ablation study of the decoder width.}
        \vspace{-0.8em}
        \tablestyle{8pt}{1.05}
        \begin{tabular}{x{20}x{24}x{24}}\toprule
        dim & NDS & mAP \\
        \midrule
        32 & \cellcolor[HTML]{efefef}\textbf{32.9} & 32.6 \\
        64 & 32.5 & 32.8 \\
        128 & \cellcolor[HTML]{efefef}\textbf{32.9} & 32.6 \\
        256 & 32.4 & \cellcolor[HTML]{efefef}\textbf{32.9} \\\bottomrule
        \end{tabular}%
	\label{tab:decoder_width}
	\end{minipage}
	\hfill
	\begin{minipage}[b]{0.48\linewidth}
	\centering
	\caption{Ablation study of the rendering technique.}
        \vspace{-0.8em}
        \tablestyle{3pt}{1.05}
        \begin{tabular}{lcc}\toprule
        Methods & NDS & mAP \\
        \midrule
        Baseline & \underline{25.2} & \underline{23.0} \\
        UniSurf~\cite{oechsle2021unisurf} & 32.5 & 32.1 \\
        VolSDF~\cite{lior2021volsdf} & 32.8 & 32.4 \\
        \rowcolor[gray]{.92}
        NeuS~\cite{wang2021neus} & \textbf{32.9\up{7.7}} & \textbf{32.6\up{9.6}} \\\bottomrule
        \end{tabular}%
	\label{tab:rendering_tech}
	\end{minipage}
\end{table}

\begin{figure}[!t]
	\centering
	\includegraphics[width=0.98\columnwidth]{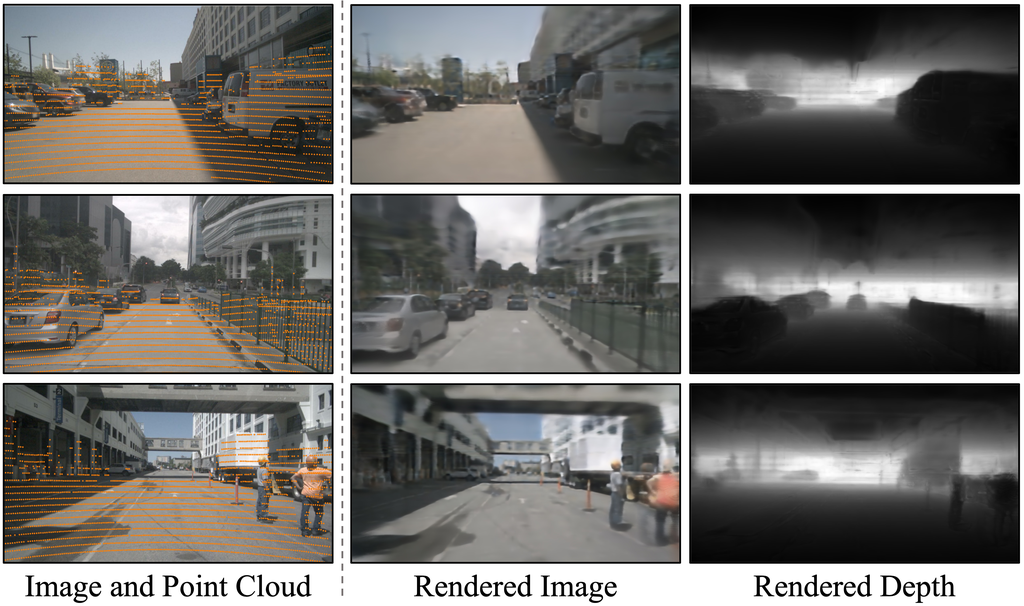}
	\vspace{-0.8em}
	\caption{\textbf{Illustration of the rendering results.} The ground truth RGB and projected point clouds, rendered RGB, and rendered depth are shown on the left, middle, and right, respectively. }
	\label{fig:render_vis}
\end{figure}

\begin{figure*}[!t]
	\centering
	\includegraphics[width=2.0\columnwidth]{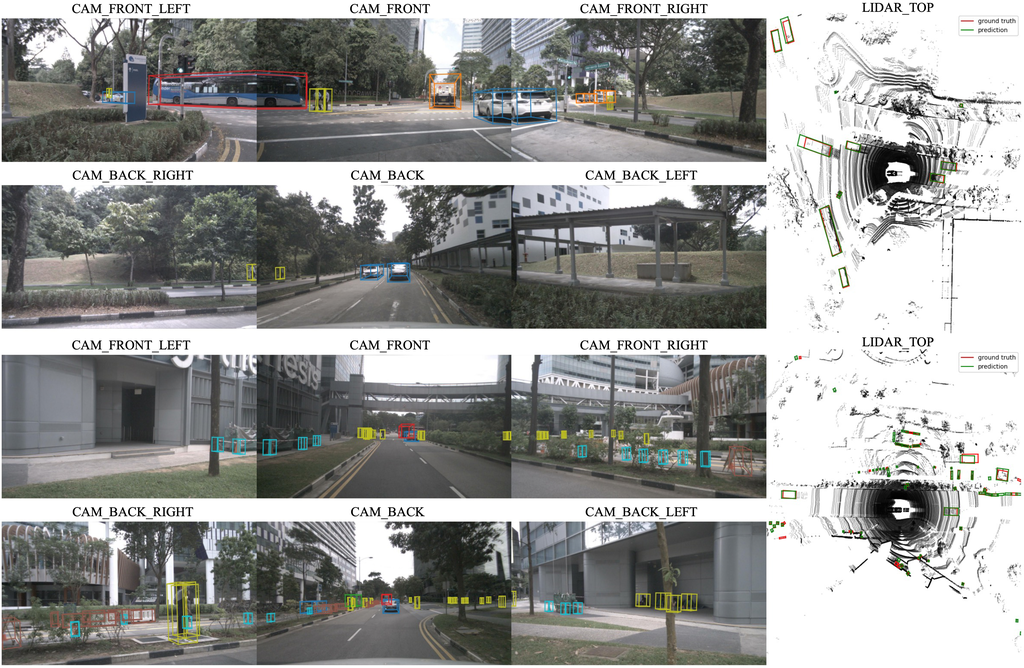}
	\vspace{-0.8em}
	\caption{\textbf{Illustration of the detection results.} The predictions are shown on multi-view images and bird's eye view with LiDAR points. }
	\label{fig:det_vis}
\end{figure*}

\subsection{Qualitative Results}
Figure~\ref{fig:render_vis} provides some qualitative results of the rendered image and depth map.
Our approach has the capability to estimate the depth of small objects, such as cars at a distance. This quality in the pre-training process indicates that our method could encode intricate and continuous geometric representations, which would benefit the downstream tasks.
In Figure~\ref{fig:det_vis}, we present 3D detection results in camera space and BEV (Bird's Eye View) space with LiDAR point clouds.
Our model can predict accurate bounding boxes for nearby objects and also shows the capability of detecting objects from far distances.

\section{Conclusions and Limitations}
\revise{
In this paper, we introduce PonderV2, a novel pre-training paradigm for 3D representation learning using differentiable neural rendering. It boosts performance across over 9 downstream tasks and achieves state-of-the-art results on 11 benchmarks, proving its flexibility and effectiveness.
}

\revise{
However, this work is an initial step and is not a 3D foundation model yet, since we have only tested on a lightweight backbone, SparseUNet. Future work should scale up datasets and backbones, and explore more tasks like reconstruction and robotic control. We aim to contribute to the development of 3D foundation models.
}

\section*{Acknowledgments}
This work was supported by the National Key R\&D Program of China (No. 2022ZD0160102). 
This work is also supported by Shanghai Artificial Intelligence Laboratory. 
This work was done during their internship at Shanghai Artificial Intelligence Laboratory by Haoyi Zhu, Honghui Yang, Xiaoyang Wu, Di Huang, Sha Zhang, and Xianglong He.

We thank the Research Support, IT, and Infrastructure team at Shanghai AI Laboratory, especially Xingpu Li, Jie Zhu, Qiang Duan, and Rui Du, for providing computational resources and network support. We also appreciate Prof. Sida Peng from Zhejiang University, Dr. Lei Bai, Dr. Xiaoshui Huang, Dr. Yuenan Hou, Mr. Jiong Wang, Mr. Kaixin Xu, Miss Chenxi Huang, Mr. Zeren Chen, Mr. Peng Ye, Mr. Shixiang Tang from Shanghai AI Laboratory, and Mr. Jiaheng Liu from Beihang University for their valuable discussions and contributions, which significantly enhanced this work.

{
\small
  \bibliographystyle{IEEEtran}
  \bibliography{main}
}

\vspace{-4em}
\begin{IEEEbiographynophoto}{Haoyi Zhu}
is a Ph.D. student at Shanghai AI Laboratory and University of Science and Technology of China. His research interests are concentrated in the areas of 3D computer vision and embodied AI.
\end{IEEEbiographynophoto}
\vspace{-4em}
\begin{IEEEbiographynophoto}{Honghui Yang}
is currently pursuing the PhD degree in the College of Computer Science and Technology at Zhejiang University. His research interests include computer vision, deep learning, and 3D object detection.
\end{IEEEbiographynophoto}
\vspace{-4em}
\begin{IEEEbiographynophoto}{Xiaoyang Wu}
received the B.S. degree from Sun Yat-sen University in 2021. He is currently working towards the Ph.D. degree in the Department of Computer Science from the University of Hong Kong. His current research interests include 3D perception and representation learning.
\end{IEEEbiographynophoto}
\vspace{-4em}
\begin{IEEEbiographynophoto}{Di Huang}
received the B.S. degree from Zhejiang University in 2020. He is currently pursuing the Ph.D. degree in the Electrical \& Information Engineering from the University of Sydney. His current research interests include 3D reconstruction and understanding.
\end{IEEEbiographynophoto}
\vspace{-4em}
\begin{IEEEbiographynophoto}{Sha Zhang}
is a Ph.D student at the University of Science and Technology of China (USTC) and currently an intern at Shanghai AI Lab at Shanghai Al Laboratory. She received Bachelor's degree from University of Science and Technology of China in 2020. Her research interests are concentrated in the areas of 3D computer vision.
\end{IEEEbiographynophoto}
\vspace{-4em}
\begin{IEEEbiographynophoto}{Xianglong He}
is currently a master candidate at Tsinghua Shenzhen International Graduate School. He received his B.S. degree from Ocean University of China. He is currently working on 3D computer vision. 
\end{IEEEbiographynophoto}
\vspace{-4em}
\begin{IEEEbiographynophoto}{Hengshuang Zhao}
is an Assistant Professor in the Department of Computer Science at The University of Hong Kong. Before that, he was a postdoctoral researcher at Massachusetts Institute of Technology and University of Oxford. He obtained his Ph.D. degree from The Chinese University of Hong Kong. His research interests line in computer vision, machine learning, and artificial intelligence. He and his team won several championships in competitive international challenges like the ImageNet Scene Parsing Challenge. He received the rising star award at the world artificial intelligence conference and was recognized as one of the most influential scholars in computer vision by AI 2000. He has served as an Area Chair for CVPR, NeurIPS, AAAI, WACV, and ACMMM. He is a member of IEEE.
\end{IEEEbiographynophoto}
\vspace{-4em}
\begin{IEEEbiographynophoto}{Chunhua Shen}
is a chair professor at Zhejiang University. He was a professor at The University of Adelaide, Australia.
\end{IEEEbiographynophoto}
\vspace{-4em}

\begin{IEEEbiographynophoto}{Yu Qiao}
(Senior Member, IEEE) is a professor with Shanghai AI Laboratory and the Shenzhen Institute of Advanced Technology (SIAT), Chinese Academy of Sciences. He has published more than 600 articles in international journals and conferences, including IEEE Transactions on Pattern Analysis and Machine Intelligence, International Journal of Computer Vision, IEEE Transactions on Image Processing, IEEE Transactions on Signal Processing, CVPR, and ICCV. His research interests include computer vision, deep learning, and bioinformation. He received the First Prize of the Guangdong Technological Invention Award, and the Jiaxi Lv Young Researcher Award from the Chinese Academy of Sciences. He is a recipient of the distinguished paper award in AAAI 2021.
\end{IEEEbiographynophoto}
\vspace{-4em}
\begin{IEEEbiographynophoto}{Tong He}
received the Ph.D. degree in computer science from the University of Adelaide, Australia in 2020. He is currently a researcher at Shanghai AI Laboratory.
\end{IEEEbiographynophoto}
\vspace{-4em}
\begin{IEEEbiographynophoto}{Wanli Ouyang} (Senior Member, IEEE) received
the Ph.D. degree from the Department of Electronic Engineering, The Chinese University of Hong Kong. His research interests include deep learning and its application to computer vision and pattern recognition, image, and video processing. He was awarded the Australian Research Council Future Fellowship, meaning that he will be exempted from teaching and can focus on research in the next four years.
\end{IEEEbiographynophoto}
\vspace{-4em}

\hfill
\newpage
{\appendices
\section{Additional Indoor Rendering Losses}
\label{appx:loss}

In our indoor experiments, we introduce additional rendering losses, as they are standard in neural construction work. However, we found that these losses do not impact downstream performance in outdoor experiments, so we chose not to apply them in that context.

{\noindent \textbf{Eikonal Regularization Loss}} $\Ls_{\text{eikonal}}$ is the widely used Eikonal loss~\cite{gropp2020implicit} for SDF regularization:
\begin{equation}
    \Ls_{\text{eikonal}} = \frac{1}{N_rN_p}\sum_{i,j}^{N_r,N_p}\left( 
\vert \nabla s(\mathbf{p}_{i,j})\vert - 1 \right)^2,
\end{equation}
where $\nabla s(\mathbf{p}_{i,j})$ denotes the gradient of SDF $s$ at location $\mathbf{p}_{i,j}$. Since SDF is a distance measure, $\Ls_{\text{eikonal}}$ encourages this distance to have a unit norm gradient at the query point.

{\noindent \textbf{Near-Surface and Free Space Loss for SDF}} Similar to iSDF~\cite{ortiz2022isdf} and GO-Surf~\cite{wang2022go}, we add additional approximate SDF supervision to help the SDF estimation. Specifically, for near-surface points, the difference between rendered depth and ground-truth depth can be viewed as the pseudo-SDF ground-truth supervision; for points far from the surface, a free space loss is used to regularize the irregular SDF value additionally. To calculate the approximate SDF supervision, we first define an indicator $b(z)$ for each sampled ray point with ray length $z$ and corresponding GT depth $D$:
\begin{equation}
    b(z) = D - z.
\end{equation}
$b(z)$ can be viewed as the approximate SDF value, which is credible only when $b(z)$ is small. Let $t$ be a human-defined threshold, which is set as $0.05$ in our paper. For sampled ray points that satisfy $b(z) \leq t$, we leverage the near-surface SDF loss to constrain the SDF prediction $s(z_{i,j})$:
\begin{equation}
    \Ls_{\text{sdf}} = \frac{1}{N_rN_p}\sum_{i,j}^{N_r,N_p} \vert s(z_{i,j}) - b(z_{i,j}) \vert.
\end{equation}
For the remaining sampled ray points, we use a free space loss:
\begin{equation}
    \Ls_{\text{free}} = \frac{1}{N_rN_p} \sum_{i,j}^{N_r,N_p} \max\left(0, e^{-\alpha\cdot s(z_{i,j})} - 1, s(z_{i,j}) - b(z_{i,j})\right),
\end{equation}
where $\alpha$ is set as 5 following the same with \cite{ortiz2022isdf,wang2022go}. Note that due to the noisy depth images, we only apply $\Ls_{\text{sdf}}$ and $\Ls_{\text{free}}$ on the rays that have valid depth values.

In our experiments, we follow a similar loss of weight with GO-Surf~\cite{wang2022go}, which set $\lambda_C$ as $10.0$, $\lambda_D$ as $1.0$, $\lambda_{\text{sdf}}$ as $10.0$ and $\lambda_{\text{free}}$ as $1.0$. We observe that the Eikonal term in our method can easily lead to over-smooth reconstructions, thus we use a small weight of $0.01$ for the EiKonal loss.

\section{Pre-Training Speed Analysis}
\label{appx:speed}

\begin{table}[!htp]\centering
\caption{\textbf{Training speed of different encoders.} The values are averaged on ScanNet semantic segmentation dataset with a batch size of $3$.}\label{tab:speed}
\tablestyle{1.75pt}{1.05}
\begin{tabular}{lccccc}\toprule
Encoder &\#Params &\#Points per Batch &GPU memory &Forward Time \\\cmidrule{1-5}
PointNet &3.53M &$\sim$40K &$\sim$6.0G &$\sim$1.82s \\
PointNet++ &0.97M &$\sim$40K &$\sim$11.0G &$\sim$2.03s \\
PointNet++ &0.97M &$\sim$85K &$\sim$21.0G &$\sim$4.08s \\
SpUNet14 &12.88M &$\sim$85K &$\sim$5.2G &$\sim$0.11s \\
SpUNet18 &23.01M &$\sim$85K &$\sim$5.2G &$\sim$0.11s \\
SpUNet34 &39.16M &$\sim$85K &$\sim$5.5G &$\sim$0.13s \\
SpUNet50 &34.03M &$\sim$85K &$\sim$9.1G &$\sim$0.14s \\
SpUNet101 &73.08M &$\sim$85K &$\sim$9.7G &$\sim$0.17s \\
\bottomrule
\end{tabular}
\end{table}
\revise{
In this paper, we use the Sparse Convolutional (SpConv) Network as the sparse encoder and pre-trained backbone for downstream tasks. To show the efficiency of SpConv, we first compare the encoder time consumed by different backbones and different encoder sizes. We test the GPU memory and forward time of different encoders on the ScanNet semantic segmentation dataset with a single NVIDIA A100 GPU and a batch size of $3$. The values are averaged in the second epoch. As shown in \tabref{tab:speed}, sparse convolution is much faster and less memory-consuming than point-based methods such as PointNet~\cite{qi2017pointnet} and PointNet++~\cite{qi2017pointnet++} even with much more model parameters. This shows its ability to be scaled up and process large datasets.}

\revise{
Moreover, we examine the training time of our rendering decoder. We test the training time on the ScanNet semantic segmentation dataset with a single NVIDIA A100 GPU and a batch size of $8$. We find that during pre-training, the total forward time is about $2.08$ seconds, and the rendering decoder takes about $1.70$ seconds which is about $81.73$\% of the total time. Although our decoder is slower than sparse encoder, the speed still remains reasonable and affordable for large-scale datasets.
}

\section{More Details on Ponder-RGBD (\secref{sec:rgbd})}
\label{appx:rgb_details}

\subsection{Back-Projecting Depth images to Point Clouds}
\label{appx:rgbd_lift}
Here we give details of the back projection process to get point clouds from depth images. Let $\mathbf{K}$ be camera intrinsic parameters, $\xi = [\mathbf{R}\vert \mathbf{t}]$ be camera extrinsic parameters, where $\mathbf{R}$ is the rotation matrix and $\mathbf{t}$ is the translation matrix. $\mathbf{X}_{uv}$ is the projected point location and $\mathbf{X}_w$ is the point location in the 3D world coordinate. Then, according to the pinhole camera model:
\begin{equation}
    s\mathbf{X}_{uv} = \mathbf{K}(\mathbf{R}\mathbf{X}_w + \mathbf{t}),
\end{equation}
where $s$ is the depth value. After expanding the $\mathbf{X}_{uv}$ and $\mathbf{X}_w$:
\begin{equation}
    s \begin{bmatrix}u\\v\\1\end{bmatrix} = \mathbf{K}(\mathbf{R} \begin{bmatrix}X\\Y\\Z\end{bmatrix} + \mathbf{t}).
\end{equation}
Then, the 3D point location can be calculated as follows:
\begin{equation}
    \begin{bmatrix}X\\Y\\Z\end{bmatrix} = \mathbf{R}^{-1}(\mathbf{K}^{-1}s\begin{bmatrix}u\\v\\1\end{bmatrix} - \mathbf{t}).
\end{equation}

\subsection{Image Synthesis From Point Clouds}
\label{appx:rgb_img}
We also validate the effectiveness of our method on another low-level task of image synthesis from point clouds.  
We use Point-NeRF~\cite{xv2022pointnerf} as the baseline. 
Point-NeRF uses neural 3D point clouds with associated neural features to render images. It can be used both for a generalizable setting for various scenes and a single-scene fitting setting. In our experiments, we mainly focus on the generalizable setting of Point-NeRF.
We replace the 2D image features of Point-NeRF with point features extracted by a DGCNN network. 
Following the same setting with PointNeRF, we use DTU~\cite{jensen2014large} as the evaluation dataset. DTU dataset is a multiple-view stereo dataset containing 80 scenes with paired images and camera poses. 
We transfer both the DGCNN encoder and color decoder as the weight initialization of Point-NeRF. We use PSNR as the metric for synthesized image quality evaluation.

The results are shown in \figref{fig:novel view}. 
By leveraging the pre-trained weights of our method, the image synthesis model is able to converge faster with fewer training steps and achieve better final image quality than training from scratch.

\begin{figure}[tb]
  \centering
  \includegraphics[width=1.0\linewidth]{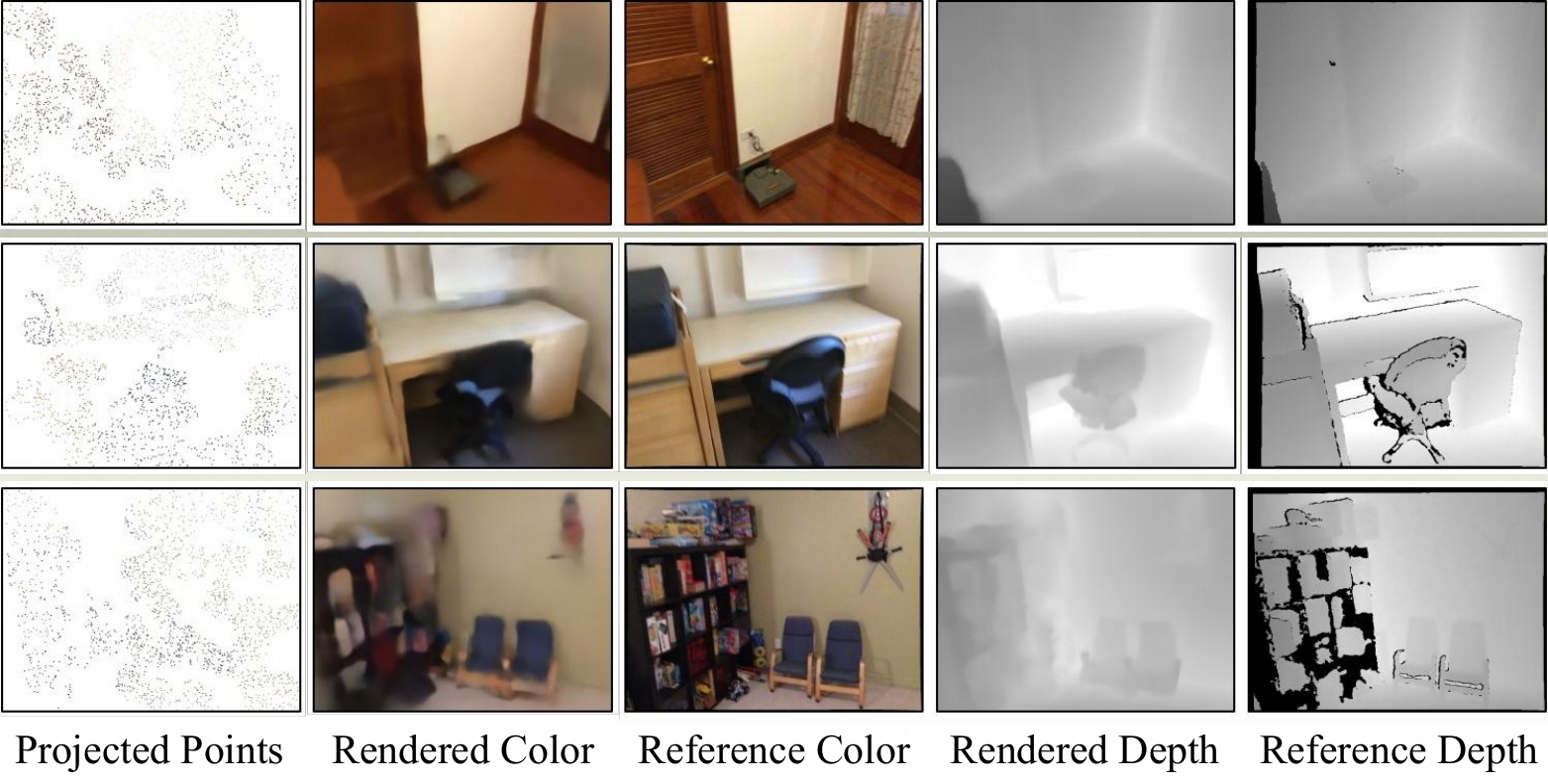}
  \vspace{-1.8em}
  \caption{\textbf{Rendered images by Ponder-RGBD} on the ScanNet validation set.
  The projected point clouds are visualized in the first column. 
  Even though input point clouds are very sparse, our model is capable of rendering color and depth images similar to the reference images. }
   \label{fig:qualitative figure}
\end{figure}
Ponder-RGBD's rendered color and depth images are shown in \figref{fig:qualitative figure}. 
As shown in the figure, even though the input point cloud is pretty sparse, our method can still render color and depth images similar to the reference images.

\begin{figure}[t]
  \centering
  \includegraphics[width=0.98\linewidth]{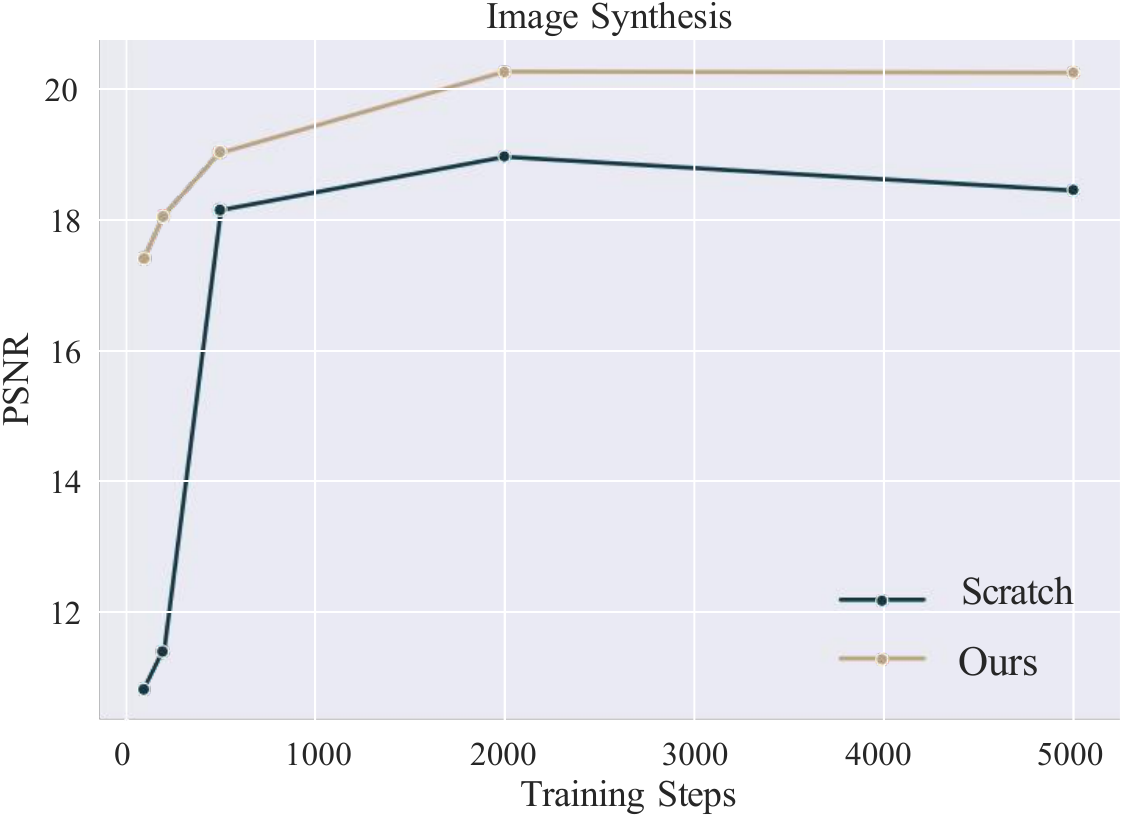}
  \vspace{-0.8em}
  \caption{\textbf{Comparison of image synthesis from point clouds.} Compared with training from scratch, 
  our Ponder-RGBD model is able to converge faster and achieve better image synthesis results.
  }
  \label{fig:novel view}
\end{figure}

\section{More Details on Indoor Study (\secref{sec:indoor})}
\label{appx:indoor_details}
\subsection{Experimental Setups}
\label{appx:indoor_setups}

In this setting, we want to pre-train a unified backbone that can be applied to various downstream tasks, whose input is directly the whole scene point clouds so that the upstream and downstream models have a unified input and encoder stage. We choose SparseUNet~\cite{choy20194d}, which is an optimized implementation of MinkUNet~\cite{choy20194d} by SpConv~\cite{spconv2022}, as $f_e^{(s)}$ due to its efficiency, whose out features $\hat{\stF}$ have $96$ channels.
We mainly focus on three widely recognized indoor datasets: ScanNet~\cite{dai2017scannet}, S3DIS~\cite{armeni2017joint} and Structured3D~\cite{zheng2020structured3d} to jointly pre-train our weights. ScanNet and S3DIS represent the most commonly used real-world datasets in the realm of 3D perception, while Structured3D is a synthetic RGB-D dataset. Given the limited data diversity available in indoor 3D vision, there exists a non-negligible domain gap between datasets, thus naive joint training may not boost performance. Point Prompt Training~\cite{wu2023ppt} (PPT) proposes to address this problem by giving each dataset its own batch norm layer. Considering its effectiveness and flexibility, we combine PPT with our universal pre-training paradigm and treat PPT as our baseline. Notably, PPT achieves state-of-the-art performance in downstream tasks with the same backbone we use, \ie, SparseUNet.

Following the pre-training phase, we discard the rendering decoder and load the encoder backbone's weights for use in downstream tasks, either with or without additional task-specific heads. Subsequently, the network undergoes fine-tuning and evaluation on each specific downstream task.
We mainly evaluate the mean intersection-over-union (mIoU) metric for semantic segmentation and mean average precision (mAP) for instance segmentation as well as object detection tasks.

\subsection{Object Reconstruction}
\label{appx:indoor_rec}

Besides high-level perception tasks, we also conduct object reconstruction experiments to see if \sexyname can work for low-level tasks. We take MCC~\cite{wu2023multiview} as our baseline and evaluate on a subset of CO3D~\cite{reizenstein2021common} dataset for fast validation. CO3D is a 3D object dataset that can be used for object-level reconstruction. Specifically, we choose $1573$ train samples and $224$ test samples from $10$ categories including \texttt{parkingmeter, baseball glove, toytrain, donut, skateboard, hotdog, frisbee, tv, sandwich} and \texttt{toybus}. We train models for 100 epochs with a learning rate of $1e^{-4}$ and an effective batch size of $64$. The problem is evaluated by the predicted occupancy of a threshold of $0.1$, the same as in MCC's original paper. The evaluation metric is F1-Score on the occupancies.
For a fair comparison, we first directly change MCC's encoder to our SparseUNet~\cite{choy20194d}. Then we average-pool the output volume into a 2D feature map and use MCC's patch embedding module to align with the shape of the decoder's input. Moreover, we adjust the shrink threshold from default $10.0$ to $3.0$, which increases the number of valid points after grid sampling. After this modification, the scratched baseline result is slightly higher than the original MCC's, as shown in \tabref{tab:indoor_objrec}. Moreover, we find that if finetuned with our pre-trained weights, it can remarkably gain nearly 2 points higher F1-Score. Note that the loaded weight is the same as previous high-level tasks, \ie it is trained on scene-level ScanNet, Structured3D, and S3DIS datasets. We directly ignore the first layer as well as the final layer of weight before finetuning, since MCC does not take normal as input and requires a different number of output channels. The results indicate that not only can our approach work well on low-level tasks, but also that the proposed method has the potential to transfer scene-level knowledge to object-level. 

\begin{table}[!tb]
\centering
\caption{\textbf{Object reconstruction} results on CO3D dataset.}
\label{tab:indoor_objrec}
\vspace{-0.8em}
\tablestyle{39pt}{1.05}
\begin{tabular}{lcc}\toprule
&F1-Score \\\cmidrule{1-2}
MCC~\cite{wu2023multiview} &\underline{63.5} \\
\rowcolor[gray]{.96}
+ SparseUNet~\cite{choy20194d} &63.9 \\
\rowcolor[gray]{.92}
+ \textbf{\sexyname} &\textbf{65.6\up{2.1}} \\
\bottomrule
\end{tabular}
\end{table}

\section{More Details on Outdoor Study (\secref{sec:outdoor})}
\label{appx:outdoor_details}

As shown in \figref{fig:improve}, \sexyname achieves significant improvements across various 3D outdoor tasks of different input modalities, which further prove the universal effectiveness of the proposed methodology.

\begin{figure}[!t]
    \centering
    \includegraphics[width=0.95\columnwidth]{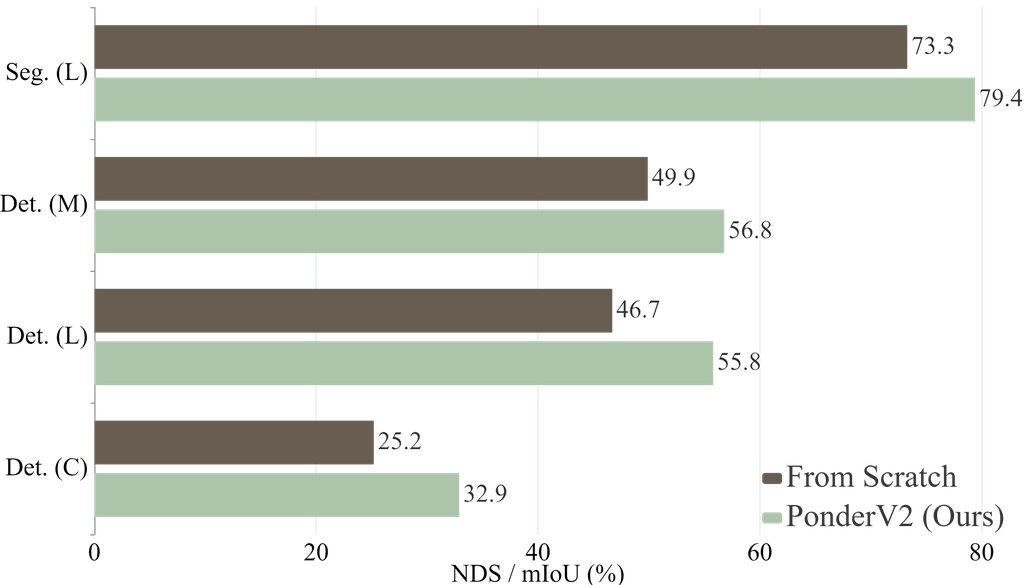}
    \vspace{-0.8em}
    \caption{\textbf{Effect of our pre-training for 3D outdoor detection and segmentation}, where C, L, and M denote camera, LiDAR, and fusion modality, respectively.}
    \label{fig:improve}
    \vspace{-1em}
\end{figure}

\subsection{Comparisons with Pre-training Methods}
\label{appx:outdoor_pre}

\noindent \textbf{Camera-based Pre-training}
In Table~\ref{tab:image_pre_train}, we conduct comparisons between \sexyname and several camera-based pre-training approaches:
1) Depth Estimator: we follow \cite{park2021dd3d} to inject 3D priors into 2D learned features via depth estimation;
2) Detector: the image encoder is initialized using pre-trained weights from MaskRCNN~\cite{he2017mask} on the nuImages dataset~\cite{caesar2020nuscenes};
3) 3D Detector: we use the weights from the widely used monocular 3D detector~\cite{wang2021fcos3d} for model initialization, which relies on 3D labels for supervision.
\sexyname demonstrates superior knowledge transfer capabilities compared to previous unsupervised or supervised pre-training methods, showing the effectiveness of our rendering-based pretext task.

\noindent \textbf{Point-based Pre-training}
For point modality, we also present comparisons with recently proposed self-supervised methods in Table~\ref{tab:point_pre_train}:
1) Occupancy-based: we implement ALSO~\cite{boulch2023also} in our framework to train the point encoder;
2) MAE-based: the leading-performing method~\cite{yang2023gd-mae} is adopted, which reconstructs masked point clouds using the chamfer distance.
3) Contrast-based: \cite{liu2021ppkt} is used for comparisons, which employs pixel-to-point contrastive learning to integrate 2D knowledge into 3D points.
Among these methods, \sexyname achieves the best NDS performance.
While \sexyname has a slightly lower mAP compared to the contrast-based method, it avoids the need for complex positive-negative sample assignments in contrastive learning.

\subsubsection{Effectiveness on Various Backbones}
\label{appx:outdoor_eff}

\noindent \textbf{Different View Transformations}
In Table~\ref{tab:view_trans}, we investigate different view transformation strategies for converting 2D features into 3D space, including BEVDet~\cite{huang2021bevdet}, BEVDepth~\cite{li2023bevdepth}, and BEVformer~\cite{li2022bevformer}.
Improvements ranging from 5.2 to 6.3 NDS are observed across different transform techniques, demonstrating the strong generalization ability of the proposed method.

\noindent \textbf{Scaling up Backbones}
To test \sexyname across different backbone scales, we adopt an off-the-shelf model, ConvNeXt, and its variants with different numbers of learnable parameters.
As shown in Table~\ref{tab:scale_up_back}, one can observe that with our \sexyname pre-training, all baselines are improved by large margins of
+6.0$\sim$7.7 NDS and +8.2$\sim$10.3 mAP.
The steady gains suggest that \sexyname has the potential to boost various state-of-the-art networks.

\begin{table}[!t]
	\centering
	\caption{Pre-training effectiveness on different view transformation strategies.}
	\vspace{-0.8em}
        \tablestyle{11.5pt}{1.05}
        \begin{tabular}{lccc}
		\toprule
		Methods & View Transform & NDS & mAP \\
		\midrule
			BEVDet & Pooling & 27.1 & 24.6 \\
                \rowcolor[gray]{.92}
			\textbf{+ \sexyname} & Pooling & \textbf{32.7\up{5.6}} & \textbf{32.8\up{8.2}} \\
			\midrule
			BEVDepth & Pooling \& Depth & 28.9 & 28.1 \\
                \rowcolor[gray]{.92}
			\textbf{+ \sexyname} & Pooling \& Depth & \textbf{34.1\up{5.2}} & \textbf{33.9\up{5.8}} \\
			\midrule
			BEVFormer & Transformer & 26.8 & 24.5  \\
                \rowcolor[gray]{.92}
			\textbf{+ \sexyname} & Transformer & \textbf{33.1\up{6.3}} & \textbf{31.9\up{7.4}} \\
		\bottomrule 
	\end{tabular}%
	\label{tab:view_trans}
\end{table}

\begin{table}[!b]
	\centering
	\caption{Pre-training effectiveness on different backbone scales.}
	\vspace{-0.8em}
        \tablestyle{2pt}{1.05}
        \begin{tabular}{lccc}
			\toprule
			\multirow{2}{*}{Methods} & \multicolumn{3}{c}{Backbone} \\\cmidrule(lr){2-4}
   & ConvNeXt-S & ConvNeXt-B & ConvNeXt-L \\
   \midrule
                UVTR-C & 25.2 / 23.0 & 26.9 / 24.4 & 29.1 / 27.7 \\
                \rowcolor[gray]{.92}
                \textbf{+\sexyname} & \textbf{32.9\up{7.7} / 32.6\up{9.6}} & \textbf{34.1\up{7.2} / 34.7\up{10.3}} & \textbf{35.1\up{6.0} / 35.9\up{8.2}} \\
			\bottomrule 
	\end{tabular}%
	\label{tab:scale_up_back}
\end{table}

}

\end{document}